\documentclass[letterpaper]{article}
\usepackage{aaai2027}
\usepackage[hyphens]{url}
\usepackage{graphicx}
\usepackage{natbib}
\usepackage{caption}
\usepackage{algorithm}
\usepackage{algorithmic}

\usepackage{newfloat}
\usepackage{listings}
\DeclareCaptionStyle{ruled}{labelfont=normalfont,labelsep=colon,strut=off}
\floatstyle{ruled}
\newfloat{listing}{tb}{lst}{}
\floatname{listing}{Listing}

\usepackage{booktabs}

\nocopyright
\usepackage{amsmath,amssymb}
\usepackage{multirow}
\usepackage{paralist}
\usepackage{tikz}
\usetikzlibrary{positioning, arrows.meta, shadows, shapes.geometric, calc, fit, backgrounds}

\newcommand{\ENNs}{\ensuremath{\mathrm{ENNs}}}

\newcommand{\sigstar}{\textsuperscript{$\ast$}}
\newcommand{\sigdag}{\textsuperscript{$\dagger$}}

\title{SALT-GNN: Handling Dense Neighborhoods in Anti-Money Laundering Graphs via Statistics-Aware Attention}

\author{
    Lidia Losavio,
    Francesco Sovrano,
    Dario Fenoglio,
    Martin Gjoreski,
    Marc Langheinrich
}
\affiliations{
    Università della Svizzera italiana (USI)\\
    Lugano, Switzerland
}
\begin{document}

\maketitle
\thispagestyle{plain} 
\pagestyle{plain}     
\begin{abstract}
Money laundering threatens financial stability and exposes institutions to penalties, motivating automated detection. Because laundering schemes often emerge through relational patterns, graph neural networks (GNNs) are increasingly used for anti-money laundering (AML). Yet AML GNNs are typically evaluated with aggregate metrics such as overall F1 score, which hide an operational issue: high-activity recipient accounts concentrate many incoming transactions, making suspicious signals harder to isolate and costlier to investigate.
We introduce a recipient-degree stratified evaluation that reports standard AML metrics across recipient-context density. Across three datasets (HI-Small, HI-Medium, and AMLSim-32k-5\%), the evaluation reveals consistent degradation in dense recipient contexts, which we trace to three GNN characteristics: the first two are known GNN limitations that AML amplifies, i.e., (1) multiset non-discriminability and (2) cardinality blindness; and (3) an attention-specific effect, i.e., in dense neighborhoods, normalized attention attenuates weak but pattern-relevant multi-hop signals.
Guided by this diagnosis, we propose SALT-GNN, a lightweight statistics-aware architecture that fuses degree-aware statistical aggregation with attention at each message-passing layer. This layer-wise fusion allows distributional and cardinality information to shape the node states used by subsequent attention steps. Ablation results support fusion placement as a key factor in dense-context performance. On HI-Small and HI-Medium it uses up to 77\% fewer parameters than task-specific graph-transformer baselines while improving dense-context F1 score by 3--6 points; on AMLSim-32k-5\%, it improves highest-degree F1 score by 16--20 points. The gains hold for both Transformer- and GAT-style attention, indicating that the benefit comes from where statistical and attentional evidence is fused rather than from a specific attention operator.
\end{abstract}

\begin{figure}[t]
  \centering
  \includegraphics[width=\linewidth]{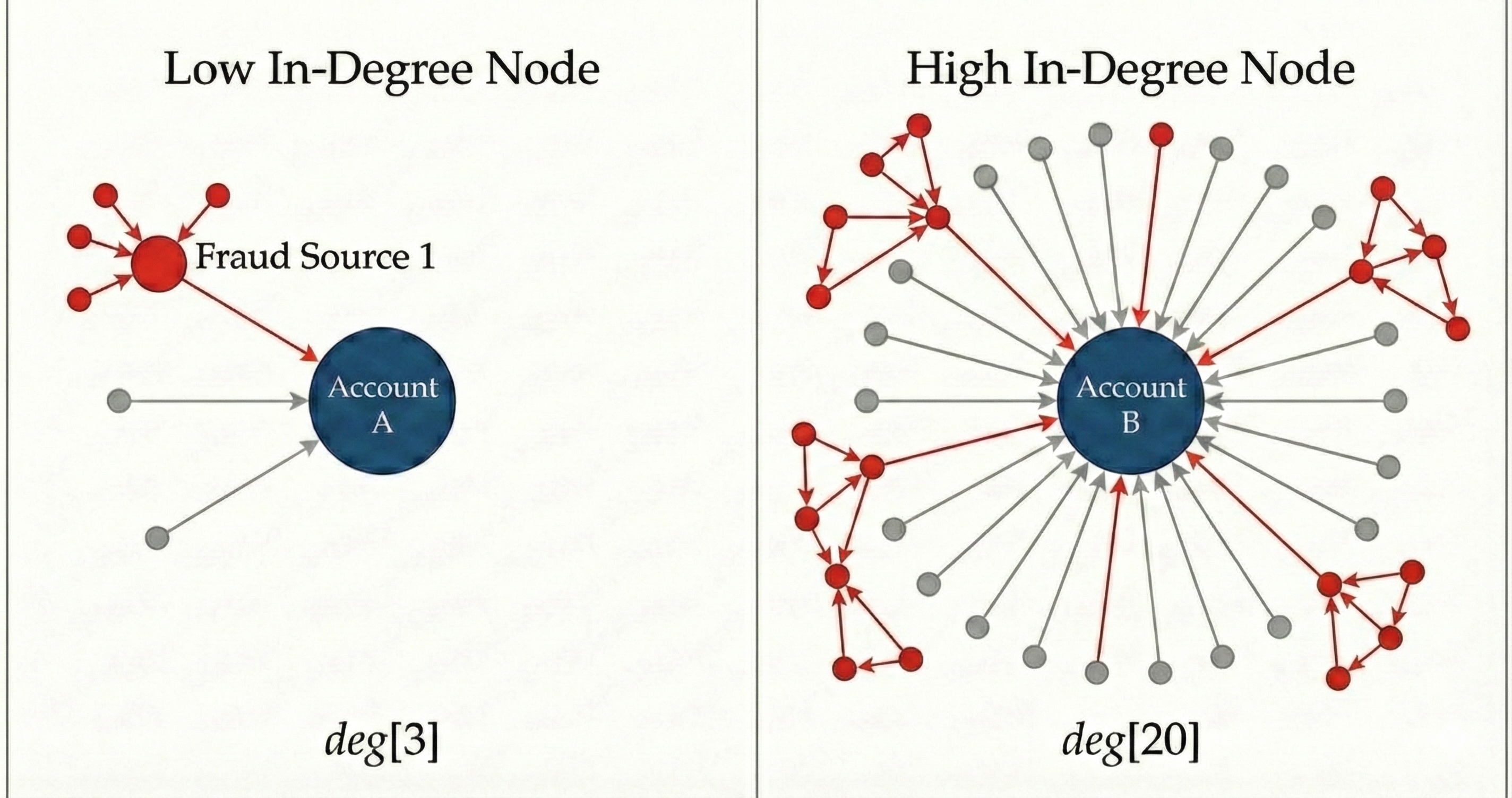}
  \caption{Recipient-context example. A suspicious incoming transaction (red) is easier to isolate at low in-degree (left) than among many normal transfers at high in-degree (right).}
  \label{fig:degree}
\end{figure}

\section{Introduction}
Money laundering conceals illicit asset origins through placement, layering, and integration, involving an estimated 2--5\% of global GDP (\$800 billion to \$2 trillion annually) \cite{UNODC_ML_estimates, AML_EU}. Because compliance failures expose financial institutions to substantial penalties, effective anti-money laundering (AML) systems are operationally essential \cite{AMLlitRev}.
A central difficulty is that laundering rarely appears as a single anomalous transaction \cite{IBMdatasets}. Instead, risk signals often emerge from relational structures such as cycles, fan-in/fan-out, and layered flows (Appendix~A). AML data is therefore naturally graph-structured, with accounts as nodes and transactions as directed edges. Graph Neural Networks (GNNs), which propagate and aggregate information over transaction graphs, have consequently become a prominent modeling family in AML research \cite{IBMdatasets,multiGNNproofs}. Although engineered-feature pipelines remain competitive \cite{GraphFeaturesPreProcessor}, recent AML-tailored GNNs provide strong benchmarks for both transaction-level edge classification and account-level node classification \cite{AMLSim,fraudGT}. However, current AML-GNN evaluation remains incomplete. Aggregate metrics such as F1 score and PR-AUC indicate whether a model performs well on average, but they do not show whether it remains reliable in the recipient contexts where many incoming transfers compete to be summarized. We call the incoming neighborhood of a receiver a recipient context. For edge classification, a transaction inherits the context of its receiver, since a transaction has no intrinsic degree; for node classification, the context is the evaluated account itself. This unit is operationally meaningful: errors in dense recipient contexts are costly, because false positives can trigger large manual reviews, while false negatives may miss collection or mixing points associated with large-scale laundering \cite{fraudDetectionLimits,Cost_of_Fraud_Prediction_Errors}.

Our central observation is that dense recipient contexts are a missing unit of AML-GNN evaluation and design. We expose them with a recipient-degree diagnostic: a post-hoc stratification of test predictions by recipient in-degree (Fig.~\ref{fig:degree}) without changing training or model selection. Across three widely used AML benchmarks—HI-Small, HI-Medium \cite{IBMdatasets}, and AMLSim-32k-5\% \cite{AMLSim,GAMLNet}—recipient-degree stratified evaluation reveals a pattern hidden by aggregate metrics: GNN performance deteriorates precisely in high in-degree recipient contexts. These are exactly the regions where detection is technically difficult and operationally most costly. This degradation is not incidental to a single model or dataset, and our results show that it is not resolved simply by increasing attention-model capacity or tuning. At the same time, the finding does not imply that attention should be abandoned. Attention remains valuable in AML because it can selectively focus on relevant neighbors among large volumes of background activity. The limitation is that, as recipient density grows, attention becomes fragile: weak but pattern-relevant multi-hop signals can be attenuated before the full laundering pattern is assembled. Thus, recipient-degree stratification localizes the failure, but it also points to a broader design question: what information should be available to the model before each attention step?

To answer this question, we map the dense-recipient failure to three aggregation limitations that AML graphs amplify: (i) limited discrimination of neighborhood multisets under single summaries \cite{PNA}, (ii) loss of cardinality information in normalized or order-statistic aggregators \cite{cardinalityProblem}, and (iii) attenuation of weak multi-hop evidence under normalized attention. These limitations are not specific to AML \cite{wu2019net,wu2022nodeformer, sun2026relieving, GIN}, but they become especially consequential in transaction graphs, where suspicious evidence is often relational, sparse within large volumes of benign activity, and meaningful only when multiple weak signals are composed into a broader pattern.
This leads to a simple design principle: attention provides useful selectivity, but dense recipient contexts require distributional and cardinality information to be incorporated into the node state before subsequent attention layers operate. 
If this information is introduced only at prediction time \cite{PNAGMDA}, later attention steps still select over representations that may not contain the cues needed to distinguish dense suspicious contexts. Our ablations support this view (\S\ref{sec:results}; Appendix~M), showing that the placement of statistical-attention fusion, rather than the mere presence of both branches, is critical for dense-region performance. 

We instantiate this principle as \textsc{SALT-GNN}, a lightweight architecture that couples degree-aware statistical aggregation with learned attention at each message-passing layer, so both distributional evidence and selective attention are available before the next propagation step. To test whether the principle depends on a particular attention operator, we instantiate it with both Transformer-style and GAT-style attention (\emph{SALT-Trans}, \emph{SALT-GAT}). Empirically, SALT improves aggregate and high-degree F1 score across all three benchmarks, using up to 77\% fewer parameters than the strongest baselines on HI-Small and HI-Medium and reaching its largest gains in the dense recipient contexts the stratified evaluation identifies. The gains hold for both backbones, supporting that the benefit comes from where evidence is fused rather than from a specific attention operator.

\paragraph{Contributions.}
\begin{compactenum}
\item \textbf{A recipient-degree diagnostic for AML-GNN evaluation.} We introduce a post-hoc protocol that reports standard AML metrics across destination in-degree ranges, revealing systematic degradation in dense recipient contexts that aggregate metrics conceal across both edge- and node-level tasks.
\item \textbf{A dense-context diagnosis for AML GNNs.} We connect the observed degradation to aggregation limitations that AML graphs amplify: multiset non-discriminability, cardinality blindness, and attenuation of weak multi-hop evidence under normalized attention.
\item \textbf{SALT-GNN, a lightweight statistics-aware attention architecture.} We propose a layer-wise fusion architecture that combines degree-aware statistical aggregation with learned attention before each propagation step, improving performance precisely in the dense regions isolated by the diagnostic.
\end{compactenum}

\noindent We release code to reproduce the stratified protocol, baselines, and ablations, with the goal of making dense-context reliability a standard part of AML-GNN evaluation.

\section{Background and Problem Formulation}
We formalize financial transaction graphs and the edge- and node-level AML tasks, review message passing and attention in standard GNNs, and introduce three general aggregation limitations, linking them to AML transaction graphs.

\subsection{Financial Transaction Graphs}

Financial transaction networks are directed multigraphs $\mathcal{G}=(\mathcal{V},\mathcal{E})$, with accounts as nodes and attributed transactions as directed edges; parallel edges encode repeated transfers.

\paragraph{Task Formulation.} We study transaction-level edge classification and account-level node classification, predicting whether a transaction or account is suspicious. This dual formulation lets us validate architectural insights across task types.

\subsection{Graph Neural Networks}

Message Passing Neural Networks (MPNNs) update node representations as
$
\mathbf{h}_v^{(k)}\!=\!\mathrm{U}^{(k)}\!\left(\mathbf{h}_v^{(k-1)}\!, \mathrm{AGG}^{(k)}\!\left(\{\mathbf{h}_u^{(k-1)}\!:\! u \in \mathcal{N}(v)\}\right)\right)
$,
where $\mathcal{N}(v)$ is the neighborhood of $v$, $\mathrm{AGG}$ is a permutation-invariant function, and $\mathrm{U}$ is typically a Multilayer Perceptron (MLP). After $k$ rounds, $\mathbf{h}_v^{(k)}$ incorporates information from up to $k$ hops. Architectures mainly differ in $\mathrm{AGG}$: GCN uses degree-normalized averaging \citep{GCN}, GAT learns attention weights \citep{GAT}, GIN uses sum aggregation \citep{GIN,WL_note}, and PNA combines multiple statistics with degree-based scaling \citep{PNA}.
Attention-based GNNs instead compute an adaptive weighted mean of neighbor representations as their aggregation function:
{\small
\begin{equation}
\label{eq:attention}
h'_i = \sum_{j\in \mathcal{N}(i)} \alpha_{ij} h_j \quad \text{s.t.} \quad \sum_{j\in \mathcal{N}(i)} \alpha_{ij} = 1
\end{equation}
}
where $h'_i$ is the updated embedding of node $i$, $h_j$ is the embedding of a neighbor $j \in \mathcal{N}(i)$, and $\alpha_{ij} \in [0,1]$ is a learned importance coefficient (typically obtained via a softmax) that distributes unit mass across all neighbors of $i$. For hub nodes with many incoming transactions, this normalization can down-weight even moderately informative neighbors to near-zero---underexplored yet problematic in pattern-centric financial graphs, where weak but coordinated multi-hop signals matter.

\subsection{Characterization of Aggregation Limitations}
\label{sec:theory}

The three characteristics below are established properties of GNN aggregation: the first two are general limitations of common neighborhood summaries~\citep{PNA, cardinalityProblem}, and the third follows from how normalized attention propagates across hops. They have not, to our knowledge, been explicitly examined in the AML setting, yet this is a regime where they matter especially, because the structure of laundering interacts directly with each of them: illicit flows hide behind benign-looking summaries, concentrate at high-volume accounts, and are deliberately arranged to look locally legitimate. In the rest of this section we therefore map each characteristic to a concrete AML reading.

\paragraph{Characteristic~1 (Multiset non-discriminability) \citep{PNA}.}\label{par:char1} A single neighborhood statistic such as the mean, maximum, minimum, or standard deviation cannot uniquely characterize a multiset, so distinct neighborhoods may collapse to similar summaries even when their risk profiles differ. This matters in AML because the collision can hide structured behavior behind a benign-looking summary: a smurfing pattern, in which an illicit total is split across many small incoming transfers, can present an average incoming amount close to that of an ordinary account while differing sharply in dispersion or extremes. 
\paragraph{Characteristic~2 (Cardinality blindness) \citep{cardinalityProblem}.}\label{par:char2} Common neighborhood summaries are normalized or order-based: mean, min, max, and standard deviation do not by themselves encode how many neighbors produced the summary. Normalized (softmax) attention likewise outputs a convex combination whose result does not expose neighborhood size, even though the attention weights are themselves shaped by it (Characteristic~3). Sum aggregation does scale with cardinality, but it entangles this size signal with feature content and, in dense neighborhoods, with large volumes of background traffic. In AML this is a direct loss of signal rather than a generic limitation: the number of incoming transfers is itself informative, since high in-degree accounts are candidate collection or mixing points where laundered flows concentrate, and cardinality also affects modeling difficulty and operational risk.
\paragraph{Characteristic~3 (Attention-induced multi-hop signal attenuation).}\label{par:char3} The normalized attention of Eq.~\eqref{eq:attention} is selective, which is valuable in AML because many transactions are irrelevant background activity. The same mechanism, however, can have a downside in AML settings: laundering is constructed through layering so that each individual transfer looks locally legitimate, which is precisely the regime in which a locally weak edge can be assigned a very small weight despite being part of a larger multi-hop pattern. Because the influence of an early-hop neighbor on a downstream node compounds multiplicatively through the attention weights along the connecting path, repeated down-weighting across successive dense hubs can attenuate such evidence before the full pattern is assembled. This is distinct from the usual benefit of attention against noise: the mechanism that filters irrelevant neighbors is the same one that can suppress weak but coordinated signals that an adversary has deliberately shaped to appear benign.

\paragraph{Implications for AML architectures.}
PNA-type models address the first two characteristics by combining multiple statistics with degree scalers, but they still aggregate all neighbors and may be sensitive to dense background traffic. Attention-based GNNs and graph transformers provide selectivity, but normalized attention remains exposed to cardinality effects and can attenuate weak multi-hop evidence. 

\section{From Diagnosis to a Design Principle: SALT-GNN}
\label{sec:salt_method}
\begin{figure}[t]
\centering
    \begin{tikzpicture}[
        scale=0.62, transform shape,
        >=latex,
        node distance=1.0cm and 1.5cm,
        box/.style={
            draw=black!60,
            rounded corners=2pt,
            thick,
            fill=white,
            drop shadow={opacity=0.15},
            minimum width=2.5cm,
            minimum height=2.5cm,
            font=\sffamily\small,
            align=center
        },
        innerbox/.style={
            draw=black!40,
            dashed,
            rounded corners=2pt,
            fill=gray!5,
            minimum width=1.8cm,
            minimum height=1.5cm
        },
        captionnode/.style={
            font=\sffamily\bfseries\footnotesize,
            anchor=south,
            yshift=2pt
        },
        arrow/.style={
            ->,
            very thick,
            color=black!70,
            rounded corners=3pt
        },
        global residual arrow/.style={
            ->,
            ultra thick,
            dashed,
            color=gray,
            rounded corners=5pt
        }
    ]
    \node[box, label={[captionnode]above left:Input}] (input) {};
    \begin{scope}[shift={(input.center)}, scale=0.6]
        \node[circle, fill=black!80, inner sep=2pt] (c) at (0,0) {};
        \foreach \ang/\n in {0/1, 72/2, 144/3, 216/4, 288/5} {
            \node[circle, fill=gray!50, inner sep=1.5pt] (n\n) at (\ang:1.0) {};
            \draw[thick, gray] (c) -- (n\n);
            \draw[thin, gray!40] (n\n) -- (0, -1.3);
        }
    \end{scope}
    \node[box, right=2.0cm of input, yshift=1.6cm, label={[captionnode]above:Statistical Channel}] (pna) {};
    \begin{scope}[shift={(pna.center)}]
        \node[innerbox] (pnainner) {};
        \node[font=\bfseries\normalsize, align=center] at (0,0.3) {$\mu, \sigma$};
        \node[font=\bfseries\normalsize, align=center] at (0,-0.3) {$\min, \max$};
    \end{scope}
    \node[box, right=2.0cm of input, yshift=-1.6cm, label={[captionnode, yshift=-18pt]below:Attention Channel}] (trans) {};
    \begin{scope}[shift={(trans.center)}, scale=0.6]
        \node[circle, fill=black!80, inner sep=2pt] (tc) at (0,0) {};
        \foreach \ang/\val in {0/2, 72/0.5, 144/1.5, 216/0.5, 288/0.5} {
            \pgfmathsetmacro{\mycolor}{\val > 1.0 ? "orange!60" : "gray!30"}
            \pgfmathsetmacro{\mysize}{\val > 1.0 ? 2.5 : 1.5}
            \node[circle, fill=\mycolor, inner sep=\mysize pt] (tn\ang) at (\ang:1.0) {};
            \draw[line width=\val pt, \mycolor] (tc) -- (tn\ang);
        }
    \end{scope}
    \path (pna.east) -- (trans.east) coordinate[midway] (midwayfusion);
    \node[
        draw=black!70,
        rounded corners,
        fill=blue!5,
        minimum width=1.2cm,
        minimum height=5.5cm,
        right=1.0cm of midwayfusion,
        label={[captionnode]above:Fusion}
    ] (fusion) {};
    \node[rotate=90, font=\sffamily\bfseries\small] at (fusion.center) {Concat $\to$ MLP};
    \node[box, right=0.8cm of fusion, minimum width=2cm, minimum height=2cm, label={[captionnode] above right:Output}] (output) {};
    \begin{scope}[shift={(output.center)}]
        \node[draw, circle, thick, fill=green!10, minimum size=1.0cm] {$\mathbf{h}_v^{(k)}$};
    \end{scope}
    \draw[arrow] (input.east) -- ++(0.4,0) |- (pna.west);
    \draw[arrow] (input.east) -- ++(0.4,0) |- (trans.west);
    \draw[arrow] (pna.east) -- node[midway, above, font=\small\sffamily, text=gray!60!black] {$\times\, s(d_v)$} (fusion.west |- pna.east);
    \draw[arrow] (trans.east) -- (fusion.west |- trans.east);
    \draw[arrow, dashed, color=gray] (input.east) -- node[midway, fill=white, font=\small, inner sep=1pt, yshift=5pt] {Residual $\mathbf{h}_v^{(k-1)}$} (fusion.west);
    \draw[arrow] (fusion.east) -- (output.west);
    \draw[global residual arrow]
        (input.north) .. controls +(0, 3.8) and +(0, 3.8) ..
        node[midway, fill=white, font=\normalsize, inner sep=3pt] {Global Residual}
        (output.north);
    \end{tikzpicture}
\caption{SALT-GNN architecture. Two parallel branches process each neighborhood: a \emph{statistical channel} computes degree-aware statistics (aggregators $\mu,\sigma,\min,\max$ rescaled by degree scalers $s(d_v)$) and an \emph{attention channel} applies selective attention to highlight salient neighbors. A learnable fusion block combines both with the residual state $\mathbf{h}_v^{(k-1)}$, producing $\mathbf{h}_v^{(k)}$ with a global residual connection.}
\label{fig:saltgnn}
\end{figure}
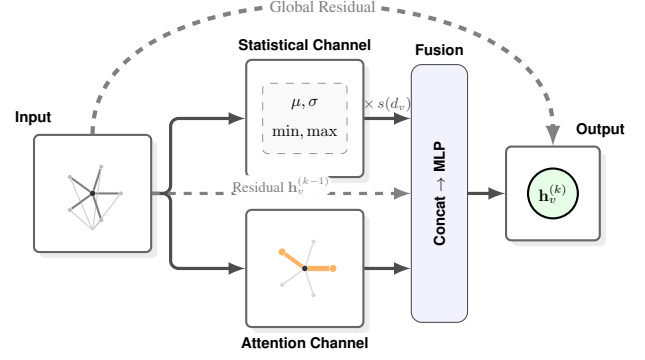

Guided by the diagnosis in \S\ref{sec:theory}, we introduce \textit{SALT-GNN}, a statistics-aware layer-wise fusion architecture for directed transaction graphs. At each message-passing layer it computes a degree-aware statistical update and an attention update, then fuses them before the next layer, so propagation uses states carrying both forms of evidence (Fig.~\ref{fig:saltgnn}).

\subsection{Layer-wise SALT update}
\label{sec:architecture}

Let $\mathcal{G}=(\mathcal{V},\mathcal{E})$ be a directed transaction multigraph, with accounts as nodes and transactions as directed edges. For a node $v$, let $\mathcal{N}_{\mathrm{in}}(v)$ denote its incoming transaction neighborhood. We write
$d_v \equiv d_{\mathrm{in}}(v) = |\mathcal{N}_{\mathrm{in}}(v)|$
for the in-degree of $v$. When multiple parallel transactions connect the same source account $u$ to the same recipient account $v$, they are treated as distinct incoming messages; for notational simplicity, we index incoming messages by $u\in\mathcal{N}_{\mathrm{in}}(v)$ and associate each message with an edge-feature vector $\mathbf{e}_{v,u}$. Let $\mathbf{h}_v^{(k-1)}\in\mathbb{R}^{d_h}$ be the representation of node $v$ entering layer $k$. Before aggregation, each incoming transaction is represented as an edge-conditioned neighbor message:
\begin{equation}
\label{eq:salt_message}
\mathbf{m}_{v,u}^{(k)}
=
\mathbf{h}_u^{(k-1)}
+
\mathbf{W}_{e}^{(k)}\mathbf{e}_{v,u},
\qquad
u\in\mathcal{N}_{\mathrm{in}}(v),
\end{equation}
where $\mathbf{W}_{e}^{(k)}$ projects edge attributes to the hidden dimension. If edge attributes are unavailable for a dataset or task, the second term is omitted. The multiset aggregated by the two SALT channels is
\[
\mathcal{M}_v^{(k)}
=
\left\{
\mathbf{m}_{v,u}^{(k)}
:
u\in\mathcal{N}_{\mathrm{in}}(v)
\right\}.
\]

\paragraph{Statistical channel.}
The statistical channel computes summaries of $\mathcal{M}_v^{(k)}$ using element-wise aggregators $\mathcal{A}=\{\mu,\sigma,\min,\max\}$ and degree scalers $\mathcal{S}$. Following PNA~\citep{PNA}, we use identity, amplification, and attenuation scalers:
\begin{equation}
\small
\label{eq:salt_scalers}
s_{\mathrm{id}}(d_v)\!=\!1,\quad \;\:
s_{\mathrm{amp}}(d_v)\!=\!\frac{\gamma_v}{\delta},\quad \;\:
s_{\mathrm{attn}}(d_v)\!=\!\frac{\delta}{\max(\gamma_v,\epsilon)},
\end{equation}
where $\gamma_v=\log(d_v+1)$, $\delta$ is the mean value of $\log(d+1)$ over the training graph, and $\epsilon>0$ is a small constant used only to avoid division by zero. The statistical summary is
\begin{equation}
\label{eq:salt_stat_summary}
\mathbf{g}_{v,\mathrm{stat}}^{(k)}
=
\bigoplus_{s\in\mathcal{S}}
\bigoplus_{\mathrm{agg}\in\mathcal{A}}
s(d_v)\cdot
\mathrm{agg}\!\left(\mathcal{M}_v^{(k)}\right),
\end{equation}
and the statistical update is
\begin{equation}
\label{eq:salt_stat_update}
\mathbf{h}_{v,\mathrm{stat}}^{(k)}
=
\mathrm{U}_{\mathrm{stat}}^{(k)}
\left(
\mathbf{h}_{v}^{(k-1)}
\,\Vert\,
\mathbf{g}_{v,\mathrm{stat}}^{(k)}
\right),
\end{equation}
where $\mathrm{U}_{\mathrm{stat}}^{(k)}$ is an MLP and $\Vert$ denotes concatenation.

\paragraph{Attention channel.}
The attention channel computes a learned weighted aggregation over the same incoming transaction messages. SALT is agnostic to the attention scoring function. For each incoming message from $u$ to $v$, let
\begin{equation}
\label{eq:salt_attention_score}
a_{v,u}^{(k)}
=
\mathrm{score}^{(k)}
\left(
\mathbf{h}_v^{(k-1)},
\mathbf{m}_{v,u}^{(k)}
\right)
\end{equation}
be an unnormalized attention score. The normalized coefficient is
\begin{equation}
\small
\label{eq:salt_attention_weights}
\alpha_{v,u}^{(k)}
=
\frac{
\exp(a_{v,u}^{(k)})
}{
\sum_{u'\in\mathcal{N}_{\mathrm{in}}(v)}
\exp(a_{v,u'}^{(k)})
},
\qquad
\sum_{u\in\mathcal{N}_{\mathrm{in}}(v)}
\alpha_{v,u}^{(k)}
=1.
\end{equation}
The attention update is then
\begin{equation}
\label{eq:salt_att_update}
\mathbf{h}_{v,\mathrm{att}}^{(k)}
=
\mathrm{U}_{\mathrm{att}}^{(k)}
\left(
\sum_{u\in\mathcal{N}_{\mathrm{in}}(v)}
\alpha_{v,u}^{(k)}
\mathbf{m}_{v,u}^{(k)}
\right),
\end{equation}
where $\mathrm{U}_{\mathrm{att}}^{(k)}$ is either an identity map or a learned projection, depending on the backbone instantiation.

We instantiate this template with two standard attention operators. In \emph{SALT-Trans}, $\mathrm{score}^{(k)}$ is a Transformer-style dot-product score~\citep{transformerConv}; in \emph{SALT-GAT}, it is a GAT-style additive score~\citep{GAT}. In both cases, the SALT contribution is not the particular attention scoring rule, but the fact that the attention update is fused layer-wise with a degree-aware statistical update.

\paragraph{Layer-wise fusion.}
The full SALT layer is the residual fusion of the previous state, the statistical update, and the attention update:
\begin{align}
\label{eq:salt_fusion_input}
\mathbf{z}_v^{(k)}
&=
\mathbf{h}_{v}^{(k-1)}
\,\Vert\,
\mathbf{h}_{v,\mathrm{stat}}^{(k)}
\,\Vert\,
\mathbf{h}_{v,\mathrm{att}}^{(k)},\\
\label{eq:salt_fusion_mlp}
\widetilde{\mathbf{h}}_v^{(k)}
&=
\mathrm{MLP}_{\mathrm{fusion}}^{(k)}
\left(
\mathbf{z}_v^{(k)}
\right),\\
\label{eq:salt_full_update}
\mathbf{h}_{v}^{(k)}
&=
\frac{1}{2}
\left(
\mathbf{h}_{v}^{(k-1)}
+
\widetilde{\mathbf{h}}_v^{(k)}
\right).
\end{align}
Eq.~\eqref{eq:salt_full_update} is the complete SALT update. The two channels do not define two independent models: after every layer, their outputs are integrated into a single node representation $\mathbf{h}_{v}^{(k)}$, which is then consumed by the next message-passing layer.

\subsection{Architectural role of the two channels}
\label{sec:salt_targeting}
Eqs.~\eqref{eq:salt_stat_summary}--\eqref{eq:salt_full_update} map directly to the aggregation characteristics of \S\ref{sec:theory}. The multi-aggregation block (Eq.~\eqref{eq:salt_stat_summary}) addresses multiset non-discriminability by exposing complementary summaries $(\mu,\sigma,\min,\max)$, and the degree scalers $s(d_v)$ address cardinality blindness by making recipient in-degree an explicit input separated from feature content. SALT does not remove softmax normalization, so the attention channel can still attenuate locally weak edges across hops; the mitigation of this effect is the coexistence of the attentive path with a parallel non-attentive statistical path inside the same layer-wise update, rather than a separate term. 


\paragraph{Propagation across hops.}
Layer-wise fusion changes what later attention layers receive as input. At layer $k+1$, messages are built from $\mathbf{h}^{(k)}$, which already contains statistical and attentional evidence from the previous hop. Thus subsequent attention scores are computed on statistics-enriched states, unlike late-fusion designs such as PNAGMDA~\citep{PNAGMDA}, where branches interact only at prediction time.

\subsection{Computational Efficiency.}
\label{sec:salt_efficiency}
Despite combining two aggregation operators, SALT-GNN remains lightweight: about 42k and 40k parameters for SALT-Trans and SALT-GAT, versus 182k for FraudGT, 61k for GAT, and 84k for a TransConv-only model. All models are instantiated at their best-validation-F1 configuration, so these differences are not tuned in SALT's favor. This compact design offers a favorable performance--parameter trade-off, which matters where computational cost is a practical constraint \citep{intesa}.

\section{Experimental Setup}
We present our stratified evaluation protocol and attention diagnostics, then datasets, baselines, and statistical testing.

\subsection{Recipient-Stratified Evaluation}
\label{sec:degstrat}

We use recipient-degree-stratified reporting as a post-hoc diagnostic: models are trained and selected under the original benchmark protocol, and only the reporting of test predictions is stratified. Alongside the stratified scores we report the overall F1 score used in prior work, making explicit how aggregate metrics conceal degradation in dense neighborhoods. Each test prediction is assigned to a degree bin according to the in-degree of its \emph{recipient context}: for node classification this is the evaluated account itself, and for edge classification it is the receiver of the transaction, since the receiver is where incoming evidence is aggregated and competing transfers must be compressed. Bins partition recipient in-degree into logarithmically increasing ranges, and within each bin we compute the same metric as in the aggregate evaluation, over the predictions whose recipient context falls in that bin (see Algorithms in Appendix~L).
\paragraph{Attention diagnostics.}
For attention-based GNNs, ENNs measures attention spread (Appendix~D), while Hit@$k$ measures fraud relevance: among nodes with at least one fraudulent incoming neighbor, it is the fraction whose top-$k$ attention weights include one. Low Hit@$k$ indicates that attention suppresses fraud-relevant evidence before multi-hop patterns can be assembled (Characteristic~3 in \S\ref{sec:theory}).

\begin{table}[hp]
    \centering
    \caption{Datasets used in our experiments.} \label{tab:amlworld_datasets}
        \small
        \setlength{\tabcolsep}{3.0pt}
        \begin{tabular}{@{}lrrrr@{}}
        \toprule
        \textbf{Dataset} & \textbf{\# Nodes} & \textbf{\# Edges} & \textbf{Illicit Rate} & \textbf{Task} \\
        \midrule
        HI-Small         & 515{,}088     & 5{,}078{,}345    & 0.102\% & Edge \\
        HI-Medium        & 2{,}077{,}023 & 31{,}898{,}238   & 0.110\% & Edge \\
        AMLSim-32k-5\%   & 32{,}000      & 757{,}044        & 5.000\% & Node \\
        \bottomrule
        \end{tabular}
\end{table}

\paragraph{Datasets.}
Because real banking data are severely restricted by privacy and regulation, AML research relies on reproducible synthetic benchmarks. We use IBM's HI-Small and HI-Medium \citep{IBMdatasets}, de facto standards for transaction-level AML evaluation \citep{tempTrag,IBMdatasets,multiGNNproofs,multiGraphAggr,GNNandEnsemble,fraudGT,GraphFeaturesPreProcessor}, and AMLSim-32k-5\% \citep{GAMLNet}, which provides account-level labels from canonical laundering patterns (Tab.~\ref{tab:amlworld_datasets}). We use strictly temporal splits for HI-Small and HI-Medium \citep{multiGNNproofs,IBMdatasets,fraudGT,tempTrag,multiGraphAggr} and the data creators' protocol for AMLSim-32k-5\% \citep{GAMLNet}.

\paragraph{Baselines.}
We instantiate SALT-GNN as SALT-Trans (Transformer-based) and SALT-GAT (GAT-based). On HI-Small and HI-Medium we compare against GIN~\citep{GIN}, PNA~\citep{PNA}, GAT~\citep{GAT}, TransConv~\citep{transformerConv}, FraudGT~\citep{fraudGT}, and PNAGMDA~\citep{PNAGMDA}, the closest late-fusion precedent. We evaluate TransConv at hidden sizes $64$ (84k params) and $128$ (318k params) to test whether dense-neighborhood degradation is merely a capacity issue \citep{di2023over}. On AMLSim-32k-5\% we also include GCN~\citep{GCN} and the AML-specialized for node classification GAMLNet~\citep{GAMLNet}. Established baselines use best validation F1 score configurations from prior work~\citep{IBMdatasets,fraudGT,GAMLNet} and tune the new ones equivalently; SALT and PNAGMDA inherit backbone hyperparameters, so gains stem from architectural fusion. Full configurations are in Appendix~I.

\paragraph{Evaluation Protocol.}
Our primary metric is positive-class F1 score, standard in AML \citep{IBMdatasets,fraudGT,multiGNNproofs, GraphFeaturesPreProcessor, GNNandEnsemble, tempTrag} to keep a single comparable metric across bins and datasets. Additionally, Appendix~F reports degree-stratified PR-AUC on HI-Small, HI-Medium, and AMLSim-32k-5\%. We report mean and standard deviation across seeds (5 for HI-Small and AMLSim-32k-5\%, 3 for HI-Medium), selecting checkpoints by best overall validation F1 score.

\paragraph{Statistical Testing.}
We test SALT against the strongest baseline and PNAGMDA on overall and densest-bin F1/PR-AUC, using exact permutation tests where possible and Welch's $t$-test for HI-Medium. Throughout the result tables, \sigstar\ and \sigdag\ mark SALT gains at $p<0.05$ and $p<0.10$ after Holm correction against the comparison baseline named in each caption; full values are in Appendix~E.
\section{Results}
\label{sec:results}

\begin{table*}[t]
\centering
\begingroup
\small
\begin{tabular}{@{}lrrrrrr@{}}
\toprule
\textbf{Model} & \textbf{[1--4]} & \textbf{[5--9]} & \textbf{[10--19]} & \textbf{[20--49]} & \textbf{[50--99]} & \textbf{overall} \\
\midrule
GIN & $0.528{\pm}0.076$ & $0.362{\pm}0.059$ & $0.479{\pm}0.124$ & $0.362{\pm}0.127$ & $0.509{\pm}0.106$ & $0.418{\pm}0.107$ \\
PNA & $0.679{\pm}0.069$ & $0.516{\pm}0.021$ & $0.662{\pm}0.015$ & $0.643{\pm}0.030$ & $0.515{\pm}0.099$ & $0.629{\pm}0.032$ \\
GAT & $0.000{\pm}0.000$ & $0.003{\pm}0.005$ & $0.002{\pm}0.003$ & $0.000{\pm}0.000$ & $0.035{\pm}0.070$ & $0.008{\pm}0.015$ \\
Trans-84k & $0.737{\pm}0.034$ & $0.542{\pm}0.034$ & $0.651{\pm}0.029$ & $0.636{\pm}0.038$ & $0.615{\pm}0.018$ & $0.645{\pm}0.031$ \\
Trans-318k & $0.756{\pm}0.013$ & $\textbf{0.574}{\pm}\textbf{0.007}$ & $0.675{\pm}0.009$ & $0.639{\pm}0.035$ & $0.581{\pm}0.053$ & $0.656{\pm}0.017$ \\
FraudGT & $0.733{\pm}0.012$ & $0.539{\pm}0.022$ & $0.674{\pm}0.031$ & $\underline{0.696{\pm}0.013}$ & $0.706{\pm}0.044$ & $0.679{\pm}0.018$ \\
PNAGMDA & $0.757{\pm}0.039$ & $0.529{\pm}0.033$ & $0.680{\pm}0.014$ & $0.682{\pm}0.020$ & $0.678{\pm}0.019$ & $0.676{\pm}0.019$ \\
\midrule
SALT-GAT & $\underline{0.765{\pm}0.027}$ & $0.557{\pm}0.021$ & $\underline{0.683{\pm}0.016}$ & $0.695{\pm}0.023$ & $\textbf{0.751}{\pm}\textbf{0.027}$\,\sigstar & $\underline{0.694{\pm}0.007}$ \\
SALT-Trans & $\textbf{0.795}{\pm}\textbf{0.028}$ & $\underline{0.573{\pm}0.042}$ & $\textbf{0.688}{\pm}\textbf{0.016}$ & $\textbf{0.723}{\pm}\textbf{0.013}$ & $\underline{0.741{\pm}0.019}$\,\sigstar & $\textbf{0.716}{\pm}\textbf{0.018}$\,\sigstar \\
\bottomrule
\end{tabular}
\caption{HI-Small F1 by recipient degree (mean $\pm$ std, 5 seeds). Bold/underline: best/second best. \sigstar: gain over PNAGMDA at $p<0.05$ (Holm-corrected permutation test). Contrasts vs.\ FraudGT are not significant on F1, though SALT-Trans significantly leads on overall PR-AUC (Appendix~E).}
\label{tab:smallhi-comprehensive}
\endgroup
\end{table*}

\begin{table*}[t]
\centering
\begingroup
\small
\setlength{\tabcolsep}{2.2pt}
\begin{tabular}{@{}lrrrrrrr@{}}
\toprule
\textbf{Model} & \textbf{[1--4]} & \textbf{[5--9]} & \textbf{[10--19]} & \textbf{[20--49]} & \textbf{[50--99]} & \textbf{[100+]} & \textbf{overall} \\
\midrule
    GIN (73k) & $0.747 \pm 0.028$ & $0.615 \pm 0.012$ & $0.672 \pm 0.020$ & $0.565 \pm 0.003$ & $0.568 \pm 0.019$ & $0.556 \pm 0.045$ & $0.608 \pm 0.011$ \\
        PNA (35k) & $0.811 \pm 0.011$ & $0.686 \pm 0.013$ & $0.722 \pm 0.013$ & $0.625 \pm 0.004$ & $0.644 \pm 0.018$ & $0.631 \pm 0.046$ & $0.671 \pm 0.006$ \\
        GAT (61k) & $0.002 \pm 0.000$ & $0.000 \pm 0.000$ & $0.000 \pm 0.000$ & $0.023 \pm 0.000$ & $0.000 \pm 0.000$ & $0.004 \pm 0.000$ & $0.003 \pm 0.001$ \\
        TransConv (84k) & $0.812 \pm 0.007$ & $0.688 \pm 0.003$ & $0.732 \pm 0.004$ & $0.641 \pm 0.005$ & $0.645 \pm 0.007$ & $0.598 \pm 0.020$ & $0.682 \pm 0.004$ \\
        TransConv (318k) & $0.766 \pm 0.011$ & $0.575 \pm 0.004$ & $0.658 \pm 0.002$ & $0.636 \pm 0.006$ & $0.647 \pm 0.027$ & $0.560 \pm 0.057$ & $0.655 \pm 0.006$ \\
        FraudGT (182k) & $0.691 \pm 0.064$ & $0.506 \pm 0.030$ & $0.591 \pm 0.041$ & $0.572 \pm 0.023$ & $0.631 \pm 0.018$ & $0.608 \pm 0.049$ & $0.615 \pm 0.017$ \\
        PNAGMDA (40k) & $0.813 \pm 0.014$ & $0.698 \pm 0.007$ & $0.718 \pm 0.006$ & $0.625 \pm 0.024$ & $0.648 \pm 0.029$ & $0.652 \pm 0.028$ & $0.675 \pm 0.020$ \\
        \midrule
        SALT-GAT (40k) & $\textbf{0.835} \pm \textbf{0.012}$ & $\textbf{0.722} \pm \textbf{0.015}$ & $\textbf{0.741} \pm \textbf{0.015}$ & $\underline{0.663 \pm 0.022}$ & $\textbf{0.683} \pm \textbf{0.021}$ & $\textbf{0.687} \pm \textbf{0.029}$\,\sigdag & $\textbf{0.708} \pm \textbf{0.019}$ \\
        SALT-Trans (42k) & $\underline{0.832 \pm 0.005}$ & $\underline{0.712 \pm 0.008}$ & $\underline{0.736 \pm 0.006}$ & $\textbf{0.665} \pm \textbf{0.008}$ & $\underline{0.677 \pm 0.008}$ & $\underline{0.673 \pm 0.018}$\,\sigstar & $\underline{0.705 \pm 0.004}$\,\sigstar \\
\bottomrule
\end{tabular}
\caption{HI-Medium F1 score by recipient degree (mean $\pm$ std, 3 seeds). Bold/underline mark best/second best. \sigstar/\sigdag\ mark SALT gains over TransConv-84k, the strongest baseline, at $p<0.05$/$p<0.10$ after Holm correction (Welch's $t$-test, $n=3$; Appendix~E). Contrasts against PNAGMDA do not reach significance after correction on this dataset (Appendix~E).}
\label{tab:mediumhi}
\endgroup
\end{table*}

\begin{table}[t]
    \centering
    \small
    \setlength{\tabcolsep}{2.0pt}
    \caption{Compact AMLSim-32k-5\% node-classification F1 score (mean $\pm$ std, 5 seeds, 3 splits). We report one sparse bin, two high-degree tail bins, and the aggregate score; the full degree-stratified table with mean $\pm$ std is in Appendix~C. \sigstar/\sigdag\ mark SALT gains over GAT, the strongest attention baseline in the extreme bin, at $p<0.05$/$p<0.10$ after Holm correction (exact paired permutation test, $n=15$); both variants also exceed PNAGMDA there at $p_{\text{Holm}}<0.001$ (Appendix~E).}
    \label{tab:amlsim}
        \begin{tabular}{@{}lrrrr@{}}
        \toprule
        \textbf{Model} & \textbf{deg[2--4]} & \textbf{deg[1000--1999]} & \textbf{deg[2000--4732]} & \textbf{overall} \\
        \midrule
        GCN        & 0.709 & 0.000 & 0.000 & 0.785 \\
        GIN        & 0.656 & 0.000 & 0.000 & 0.751 \\
        PNA        & 0.783 & 0.578 & 0.178 & 0.852 \\
        GAMLNet    & 0.745 & 0.567 & 0.000 & 0.822 \\
        TransConv  & 0.743 & 0.760 & 0.442 & 0.827 \\
        GAT        & 0.740 & 0.632 & 0.573 & 0.821 \\
        PNAGMDA    & 0.706 & 0.633 & 0.044 & 0.826 \\
        \midrule
        SALT-GAT   & \textbf{0.794} & \textbf{0.827} & \textbf{0.782}\sigstar & \textbf{0.855} \\
        SALT-Trans & \textbf{0.794} & \underline{0.811} & \underline{0.733}\sigdag & \underline{0.855} \\
        \bottomrule
        \end{tabular}
\end{table}

Tables~\ref{tab:smallhi-comprehensive}--\ref{tab:amlsim} show a consistent pattern: aggregate F1 score hides dense-recipient degradation, while SALT remains competitive overall and is strongest in the highest-degree regimes. 
We discuss the results through three research questions (RQ1--RQ3).

\textbf{RQ1: How do vanilla GNNs behave as structural complexity increases?} Across benchmarks, standard GNNs degrade as in-degree grows. On the IBM tasks (Tables~\ref{tab:smallhi-comprehensive}--\ref{tab:mediumhi}), PNA and TransConv drop in dense bins; increasing TransConv from 84k to 318k parameters lowers dense-bin F1 score, suggesting a structural rather than capacity-only limitation. GAT collapses on IBM (overall F1 score $\approx$0.008), consistent with prior AML results~\citep{IBMdatasets,multiGNNproofs,fraudGT}. FraudGT is robust on HI-Small but scales less reliably to HI-Medium, and the deg[5--9] dip is linked to unstructured ''agent-based'' fraud (Appendix~J). AMLSim (Table~\ref{tab:amlsim}) similarly shows high overall F1 score but tail degradation, supporting that high-degree hubs remain challenging across tasks.

\begin{table}[t]
    \centering
    \caption{Hit@20 on HI-Small deg[50--99] (mean $\pm$ std, 5 seeds): fraction of nodes with at least one fraudulent neighbor among the top-20 attention weights, at the first (L0) and second (L1) hop. We set $k=20$ to match TransConv's effective neighborhood in deg[50--99], which is close to $20$ at both hops ($\ENNs\approx20.7$ at L0 and $17.0$ at L1; Appendix~D).
    Higher is better: it shows that the fraud signal remains among the most-attended neighbors.}
    \label{tab:hitk}
    \begin{tabular}{@{}lcc@{}}
    \toprule
    \textbf{Model} & \textbf{Hit@20 (L0)} & \textbf{Hit@20 (L1)} \\
    \midrule
    TransConv (84k) & $0.352 \pm 0.038$ & $0.579 \pm 0.037$ \\
    SALT-GAT        & $\textbf{0.468} \pm \textbf{0.054}$ & $\underline{0.746 \pm 0.138}$ \\
    SALT-Trans      & $\underline{0.449 \pm 0.052}$ & $\textbf{0.774} \pm \textbf{0.136}$ \\
    \bottomrule
    \end{tabular}
\end{table}

\textbf{RQ2: How do attention-based GNNs behave in dense transaction graphs?}
Attention \emph{spread} alone does not explain dense-bin behavior (Appendix~D): a model can attend broadly but rank fraud-relevant edges poorly, or narrowly and miss them. We therefore use Hit@20 as the primary diagnostic. As shown in Table~\ref{tab:hitk}, in deg[50--99] TransConv reaches Hit@20 of only $0.352$ at L0, so in roughly 65\% of cases no fraudulent neighbor appears among the top-20 attention weights; once such an edge is down-weighted early, multi-hop propagation can attenuate it further (\S\ref{sec:theory}). Both SALT variants repair this: Hit@20 rises to $0.449$/$0.468$ at L0 (SALT-Trans/SALT-GAT) and from $0.579$ to $0.774$/$0.746$ at L1. The statistics-aware pathway thus does not merely reshape attention entropy; it keeps weak fraud-relevant evidence available across hops even when learned attention would suppress it. 
\textbf{RQ3: Does SALT-GNN help in extreme-degree areas and across tasks?}
On the IBM edge benchmarks, SALT leads in the densest bins: on HI-Small deg[50--99], SALT-Trans/SALT-GAT reach 0.741/0.751 vs.\ FraudGT 0.706 and TransConv 0.615; on HI-Medium deg[100+], 0.673/0.687 vs.\ the best baseline PNAGMDA at 0.652---with competitive or best overall F1 score. On the AMLSim node task the effect persists in the extreme tail: in deg[2000--4732], SALT-Trans/SALT-GAT reach 0.733/0.782 vs.\ TransConv 0.442 and GAT 0.573, with the SALT-GAT gain significant at $p<0.05$ and SALT-Trans at $p<0.10$ after correction.
This separates SALT from PNAGMDA~\citep{PNAGMDA}, which fuses the same branches only at prediction time: it trails SALT on HI-Small overall (0.676 vs.\ 0.716) and in the densest bin (0.678 vs.\ 0.751), and collapses to 0.044 in AMLSim deg[2000--4732]. Since this contrast is between-model, we also isolate fusion placement with a within-model knockout: moving SALT's fusion from per-layer to prediction-time, operators fixed, still solves the task (probe PR-AUC $0.995$) yet collapses second-hop attention routing onto the planted fraud from AUC $1.00$ to $0.50$ (chance), while per-layer SALT keeps it at $1.00$ (Appendix~M). This shows that \emph{where} the two views interact---not which branches are combined---drives the dense-region gain.
\paragraph{Characteristic-level ablation.}
We ablate SALT-Trans on HI-Small (Table~\ref{tab:ablation_char}; full F1 score/PR-AUC in Appendix~F), removing C1 (multi-aggregation) and C2 (degree-aware encoding) one at a time, then the statistical channel as a whole. C3 is isolated by toggling the statistical channel on and off at fixed attention: removing it (\emph{no statistics}) leaves attention unmitigated, while the full model mitigates it; PNA (no attention) and TransConv (unmitigated attention) bound the comparison. Dropping multi-aggregators (mean only) lowers deg[50--99] F1 score from $0.741$ to $0.636$, with std restoring most of it ($0.720$), supporting C1; removing degree-awareness gives a comparable drop ($\to0.616$) at the highest variance, supporting C2. Attention alone without the statistical channel is less reliable in the dense bin (TransConv $0.615$, PNA $0.515$), but complementing it with statistics in SALT-Trans reaches $0.741$. Tellingly, removing the statistical channel entirely (\emph{no statistics}) reduces dense-bin F1 score to $0.637$ and doubles its variance while overall F1 score barely moves ($0.698$ vs.\ $0.716$), supporting that the dense-region gain comes from the statistical channel and that aggregate F1 score hides it.

\begin{table}[t]
    \centering
    \caption{Characteristic-level ablation on HI-Small (F1 score, mean $\pm$ std, 5 seeds).
    \checkmark/$\circ$/$\times$ denote full/partial/absent.
    \textbf{C1} (multi-aggregation): $\circ$ = two aggregators, $\times$ = mean only.
    \textbf{C2} (degree-aware encoding): $\times$ = absent.
    \textbf{C3} (attention handling): \checkmark\ = attention with multi-hop mitigation (statistics preserve down-weighted signal), $\circ$ = attention without mitigation, $\times$ = no attention.
    C1, C2 are toggled internally on SALT-Trans; \emph{no statistics} removes the whole channel; PNA and TransConv are references. Full F1 score/PR-AUC version in Appendix~F.}
    \label{tab:ablation_char}
    \small
    \setlength{\tabcolsep}{3.0pt}
    \begin{tabular}{@{}lcccrr@{}}
    \toprule
    \textbf{Model} & \textbf{C1} & \textbf{C2} & \textbf{C3} & \textbf{deg[50--99]} & \textbf{overall} \\
    \midrule
    SALT-Trans (full)        & \checkmark & \checkmark & \checkmark & $0.741\pm0.019$ & $0.716\pm0.018$ \\
    \,\,$\hookrightarrow$ mean+std only & $\circ$ & \checkmark & \checkmark & $0.720\pm0.042$ & $0.712\pm0.004$ \\
    \,\,$\hookrightarrow$ mean only     & $\times$ & \checkmark & \checkmark & $0.636\pm0.065$ & $0.690\pm0.020$ \\
    \,\,$\hookrightarrow$ no degree     & \checkmark & $\times$ & \checkmark & $0.616\pm0.103$ & $0.713\pm0.013$ \\
    \,\,$\hookrightarrow$ no statistics & $\times$ & $\times$ & $\circ$ & $0.637\pm0.032$ & $0.698\pm0.016$ \\
    \midrule
    PNA \textit{(ref.)}      & \checkmark & \checkmark & $\times$ & $0.515\pm0.099$ & $0.629\pm0.032$ \\
    TransConv \textit{(ref.)} & $\times$ & $\times$ & $\circ$ & $0.615\pm0.018$ & $0.645\pm0.031$ \\
    \bottomrule
    \end{tabular}
\end{table}

\section{Discussion, Limitations, and Future Work}
\label{sec:discussion}
On HI-Small, SALT-GAT and SALT-Trans train in 140s and 142s per epoch, close to PNA (135s) and TransConv-318k (123s) and far faster than FraudGT (242s) under the same configuration (Appendix~K). SALT thus improves dense-region behavior without a heavy task-specific graph transformer---operationally relevant, since high-activity accounts concentrate investigation cost and missed-risk exposure.
Our evaluation is restricted to public synthetic AML benchmarks, which, though standard, may not reflect the full heterogeneity of products, institutions, labeling practices, and cross-border flows in deployed systems; validating recipient-degree behavior and SALT-style hybrids on proprietary, multi-institution, and temporal graphs is an important next step.

\section{Related Work}
AML and financial-fraud detection have been studied with graph feature pre-processing, GNNs, temporal models, and graph transformers. GFP~\citep{GraphFeaturesPreProcessor} extracts structural features for tree ensembles on the IBM AML benchmarks~\citep{IBMdatasets}; GAMLNet~\citep{GAMLNet}, MultiGNN-style architectures~\citep{multiGNNproofs,multiGraphAggr}, TeMP-TraG~\citep{tempTrag}, temporal GNNs for online payments~\citep{onlinePayments}, and FraudGT~\citep{fraudGT} demonstrate the value of graph and temporal structure. Other work addresses scalable semi-supervised learning and ensembles~\citep{GNNandEnsemble}, class imbalance~\citep{pickAndChoose,balancergnn}, LLM--GNN integration~\citep{flag}, reverse message passing~\citep{reverseMP}, port numbering~\citep{portNumberGNN}, ego identifiers~\citep{egoID}, and specialized pre-processing~\citep{recentPreProc}. We focus on vanilla message passing and lightweight hybrids to isolate how standard aggregation and attention behave across recipient-degree regimes. PNAGMDA~\citep{PNAGMDA} is the closest architectural precedent because it combines PNA and GAT branches; we evaluate it directly and distinguish SALT-GNN by layer-wise rather than prediction-time fusion. Dense-subgraph methods~\citep{denseGraph,denseGraph2} are complementary: they search for suspicious dense communities in an unsupervised or subgraph-centric way, whereas we use recipient-degree stratified evaluation to analyze supervised GNN behavior across graph regimes. Work on scalable fraud pipelines~\citep{trainingCosts,intesa} emphasizes production constraints; we instead show that lightweight supervised GNNs improve dense-region robustness without heavy graph transformers.

\section{Conclusion}
We introduced recipient-degree stratified evaluation for AML GNNs, showing that aggregate metrics hide systematic degradation in dense, high-activity recipient contexts, and SALT-GNN, which fuses degree-aware statistical aggregation and attention at every message-passing layer. Across three benchmarks, SALT improves dense-bin performance with a compact footprint, and our ablations and Hit@$k$ diagnostics support the central claim: layer-wise interaction between statistics and attention is more robust in dense neighborhoods than either view alone or late fusion.

\bibliography{references}

\appendix
\section{Money Laundering Known Patterns}
\label{apx:patterns}

\begin{figure}[!htbp]
  \centering
  \includegraphics[width=\linewidth]{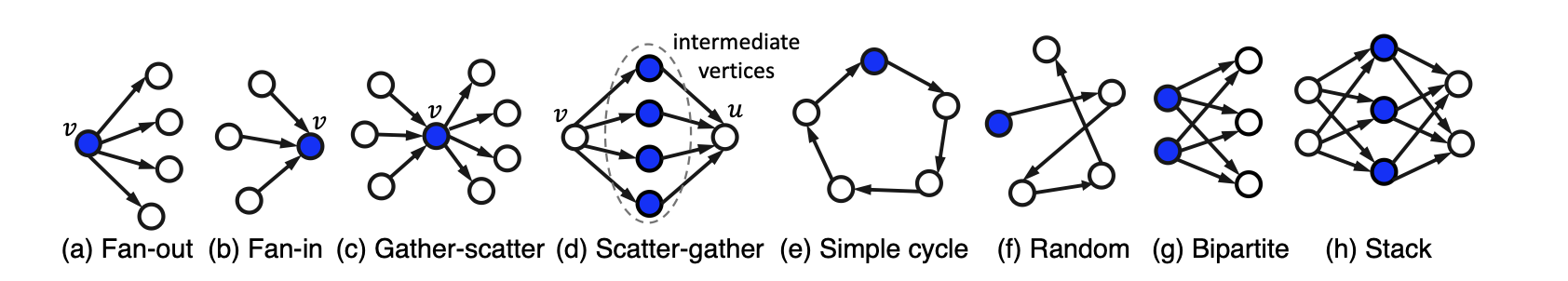}
  \caption{Canonical money laundering patterns observed in AML datasets. The blue node denotes the main laundering account; depending on the task, all nodes and transactions involved in the pattern are typically labeled as fraudulent.}
  \label{fig:patterns}
\end{figure}


\section{Degree-Stratified Evaluation}

\begin{figure}[!htbp]
  \centering
  \includegraphics[width=\linewidth]{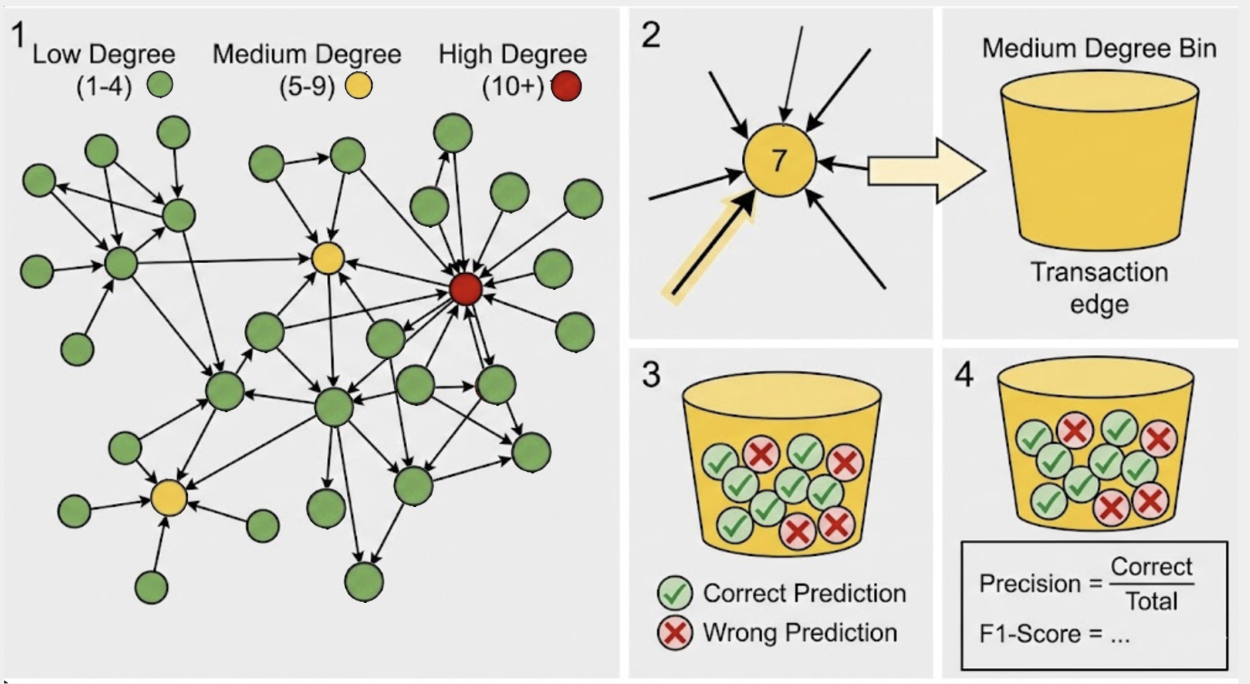}
  \caption{Degree-stratified evaluation: from an overall metric to a breakdown by the volume of incoming transactions per node (account).}
  \label{fig:degstrat}
\end{figure}

Figure~\ref{fig:degstrat} illustrates the recipient-degree stratified evaluation protocol, which reorganizes a single aggregate metric into a breakdown by the volume of incoming transactions per node (account).

\section{Full AMLSim Degree-Stratified Results}
\label{apx:amlsim_full}
The complete degree-stratified F1 results on AMLSim-32k-5\% are reported in Table~\ref{tab:amlsim_full}, presented together with the corresponding PR-AUC breakdown in the Threshold-Free Metric section below, so that the two large AMLSim tables can be read side by side.

\section{Effective Number of Neighbors Diagnostics}
\label{apx:enns}

For completeness, we report entropy-based attention-spread diagnostics. For a node $i$ at layer $\ell$, let $\boldsymbol{\alpha}_i^{(\ell)}$ be the normalized attention vector over its incoming neighbors. We compute
\begin{align*}
H(\boldsymbol{\alpha}_i^{(\ell)}) &=-\sum_j \alpha_{ij}^{(\ell)}\log\alpha_{ij}^{(\ell)}, \\
\ENNs(\boldsymbol{\alpha}_i^{(\ell)}) &=\exp\!\big(H(\boldsymbol{\alpha}_i^{(\ell)})\big).
\end{align*}
Uniform attention gives $\ENNs=d_i$, while single-neighbor concentration gives $\ENNs=1$. ENNs measures attention spread, not ranking correctness; the main paper therefore uses Hit@20 as the primary attention-preservation diagnostic.

\begin{table*}[hp]
    \centering
    \setlength{\tabcolsep}{6pt}
    \caption{Attention diagnostics on HI-Small: mean (5 seeds) Effective Number of Neighbors (\ENNs) at the first (L0) and second (L1) hop, computed as $\ENNs=\exp(H)$ with $H$ the Shannon entropy. Uniform attention gives $\ENNs=d$ (degree); single-neighbor focus gives $\ENNs=1$.}
    \label{tab:attention_diagnostics_merged}
    \small
    \begin{tabular}{@{}l r r r r r r r r@{}}
    \toprule
    \multirow{2}*{\textbf{Degree Bin}} & \multicolumn{2}{c}{\textbf{GAT}} & \multicolumn{2}{c}{\textbf{TransConv (84k)}} & \multicolumn{2}{c}{\textbf{SALT-GAT}} & \multicolumn{2}{c}{\textbf{SALT-Trans}} \\
    \cmidrule(lr){2-3} \cmidrule(lr){4-5} \cmidrule(lr){6-7} \cmidrule(lr){8-9}
    & \textbf{F1} & $\mathbf{\ENNs}$ \textbf{(L0/L1)} & \textbf{F1} & $\mathbf{\ENNs}$ \textbf{(L0/L1)} & \textbf{F1} & $\mathbf{\ENNs}$ \textbf{(L0/L1)} & \textbf{F1} & $\mathbf{\ENNs}$ \textbf{(L0/L1)} \\
    \midrule
    deg[1--1]   & $0.000$ & $1.00 / \phantom{1}1.00$ & $0.779$ & $1.00 / \phantom{1}1.00$ & $0.790$ & $1.00 / \phantom{1}1.00$ & $0.772$ & $1.00 / 1.00$ \\
    deg[2--4]   & $0.000$ & $2.89 / \phantom{1}1.78$ & $0.758$ & $2.64 / \phantom{1}2.53$ & $0.765$ & $3.03 / \phantom{1}2.30$ & $0.795$ & $2.45 / 2.10$ \\
    deg[5--9]   & $0.003$ & $6.43 / \phantom{1}3.06$ & $0.542$ & $4.89 / \phantom{1}4.22$ & $0.557$ & $6.15 / \phantom{1}3.85$ & $0.573$ & $4.62 / 2.75$ \\
    deg[10--19] & $0.002$ & $15.44 / \phantom{1}5.57$ & $0.651$ & $8.55 / \phantom{1}7.33$ & $0.683$ & $14.82 / \phantom{1}9.13$ & $0.688$ & $10.74 / 4.98$ \\
    deg[20--49] & $0.000$ & $28.76 / \phantom{1}7.92$ & $0.658$ & $13.25 / 10.06$ & $0.695$ & $28.35 / 15.31$ & $0.723$ & $19.59 / 6.95$ \\
    deg[50--99] & $0.035$ & $56.60 / 12.91$ & $0.603$ & $20.69 / 17.04$ & $0.751$ & $53.43 / 25.02$ & $0.741$ & $35.88 / 9.59$ \\
    \bottomrule
    \end{tabular}
\end{table*}

\section{Statistical Significance Tests}
\label{apx:significance}

We test whether each SALT variant outperforms the relevant baselines beyond seed variability, on two contrasts per benchmark (overall performance and the densest populated degree bin) and, for the IBM datasets, on both F1 score and PR-AUC. Besides the strongest baseline, we also test against PNAGMDA, the closest dual-branch precedent, to isolate per-layer from prediction-time fusion. The test matches each benchmark's design. On AMLSim, runs are paired by split$\times$seed (identical graph), so we use an exact paired permutation test ($n=15$; Table~\ref{tab:significance_amlsim}). On HI-Small we use an exact two-sample permutation test ($n=5$). On HI-Medium ($n=3$) the permutation null admits only $\binom{6}{3}=20$ assignments, lower-bounding the two-sided $p$ at $0.10$; we therefore use a two-sided Welch's $t$-test, which retains power at small $n$ (Table~\ref{tab:significance}). Within each family of related tests we apply Holm correction and report $p_{\text{Holm}}$; the markers \sigstar~($p<0.05$) and \sigdag~($p<0.10$) in the main tables refer to these corrected values. Given the small seed counts on the IBM datasets, we report effect sizes ($\Delta$) alongside $p$-values and treat those tests as indicative.

\begin{table*}[!htbp]
\centering
\caption{Statistical significance of SALT variants against the strongest baselines on HI-Small and HI-Medium.
We test two contrasts (overall and the densest \emph{populated} degree bin: deg[50--99] for HI-Small,
deg[100+] for HI-Medium) on F1 score and PR-AUC. HI-Small ($n=5$): exact two-sample permutation test.
HI-Medium ($n=3$): two-sided Welch's $t$-test, since the permutation null admits only
$\binom{6}{3}=20$ assignments (minimum two-sided $p=0.10$). $\Delta$ is the mean difference (SALT $-$ baseline).
$p_{\text{Holm}}$ applies Holm correction within each family of four tests
(the two contrasts $\times$ two metrics for a given SALT variant, baseline, and dataset).
\protect\sigstar~$p<0.05$, \protect\sigdag~$p<0.10$ (on $p_{\text{Holm}}$). Effect sizes are reported alongside
$p$-values; given the small seed counts, tests are indicative.}
\label{tab:significance}
\small
\begin{tabular}{@{}llccccc@{}}
\toprule
\textbf{Comparison} & \textbf{Metric} & \textbf{SALT} & \textbf{Base} & \textbf{$\Delta$} & \textbf{$p$} & \textbf{$p_{\text{Holm}}$} \\
\midrule
\multicolumn{7}{@{}l}{\textit{HI-Small} ($n=5$, exact permutation test)} \\
\midrule
SALT-GAT vs.\ FraudGT & Overall F1 & 0.694 & 0.679 & +0.015 & 0.397 & 0.397 \\
SALT-GAT vs.\ FraudGT & Overall PR-AUC & 0.685 & 0.664 & +0.021 & 0.143 & 0.333 \\
SALT-GAT vs.\ FraudGT & deg[50--99] F1 & 0.751 & 0.706 & +0.045 & 0.111 & 0.333 \\
SALT-GAT vs.\ FraudGT & deg[50--99] PR-AUC & 0.730 & 0.659 & +0.072 & 0.071 & 0.286 \\
SALT-GAT vs.\ PNAGMDA & Overall F1 & 0.694 & 0.676 & +0.018 & 0.183 & 0.365 \\
SALT-GAT vs.\ PNAGMDA & Overall PR-AUC & 0.685 & 0.668 & +0.018 & 0.302 & 0.365 \\
SALT-GAT vs.\ PNAGMDA & deg[50--99] F1 & 0.751 & 0.678 & +0.073 & 0.008 & 0.032\sigstar \\
SALT-GAT vs.\ PNAGMDA & deg[50--99] PR-AUC & 0.730 & 0.660 & +0.071 & 0.040 & 0.119 \\
\addlinespace
SALT-Trans vs.\ FraudGT & Overall F1 & 0.716 & 0.679 & +0.037 & 0.040 & 0.119 \\
SALT-Trans vs.\ FraudGT & Overall PR-AUC & 0.702 & 0.664 & +0.038 & 0.008 & 0.032\sigstar \\
SALT-Trans vs.\ FraudGT & deg[50--99] F1 & 0.741 & 0.706 & +0.035 & 0.119 & 0.119 \\
SALT-Trans vs.\ FraudGT & deg[50--99] PR-AUC & 0.726 & 0.659 & +0.067 & 0.040 & 0.119 \\
SALT-Trans vs.\ PNAGMDA & Overall F1 & 0.716 & 0.676 & +0.040 & 0.032 & 0.032\sigstar \\
SALT-Trans vs.\ PNAGMDA & Overall PR-AUC & 0.702 & 0.668 & +0.034 & 0.008 & 0.032\sigstar \\
SALT-Trans vs.\ PNAGMDA & deg[50--99] F1 & 0.741 & 0.678 & +0.063 & 0.008 & 0.032\sigstar \\
SALT-Trans vs.\ PNAGMDA & deg[50--99] PR-AUC & 0.726 & 0.660 & +0.066 & 0.016 & 0.032\sigstar \\
\midrule
\multicolumn{7}{@{}l}{\textit{HI-Medium} ($n=3$, Welch's $t$-test)} \\
\midrule
SALT-GAT vs.\ TransConv-84k & Overall F1 & 0.708 & 0.682 & +0.026 & 0.132 & 0.264 \\
SALT-GAT vs.\ TransConv-84k & Overall PR-AUC & 0.699 & 0.661 & +0.038 & 0.045 & 0.136 \\
SALT-GAT vs.\ TransConv-84k & deg[100+] F1 & 0.687 & 0.598 & +0.089 & 0.020 & 0.081\sigdag \\
SALT-GAT vs.\ TransConv-84k & deg[100+] PR-AUC & 0.684 & 0.637 & +0.047 & 0.204 & 0.264 \\
SALT-GAT vs.\ PNAGMDA & Overall F1 & 0.708 & 0.675 & +0.033 & 0.098 & 0.295 \\
SALT-GAT vs.\ PNAGMDA & Overall PR-AUC & 0.699 & 0.664 & +0.035 & 0.066 & 0.265 \\
SALT-GAT vs.\ PNAGMDA & deg[100+] F1 & 0.687 & 0.652 & +0.035 & 0.216 & 0.432 \\
SALT-GAT vs.\ PNAGMDA & deg[100+] PR-AUC & 0.684 & 0.672 & +0.012 & 0.669 & 0.669 \\
\addlinespace
SALT-Trans vs.\ TransConv-84k & Overall F1 & 0.705 & 0.682 & +0.023 & 0.006 & 0.024\sigstar \\
SALT-Trans vs.\ TransConv-84k & Overall PR-AUC & 0.699 & 0.661 & +0.038 & 0.008 & 0.024\sigstar \\
SALT-Trans vs.\ TransConv-84k & deg[100+] F1 & 0.673 & 0.598 & +0.075 & 0.021 & 0.043\sigstar \\
SALT-Trans vs.\ TransConv-84k & deg[100+] PR-AUC & 0.683 & 0.637 & +0.046 & 0.146 & 0.146 \\
SALT-Trans vs.\ PNAGMDA & Overall F1 & 0.705 & 0.675 & +0.030 & 0.088 & 0.265 \\
SALT-Trans vs.\ PNAGMDA & Overall PR-AUC & 0.699 & 0.664 & +0.035 & 0.049 & 0.195 \\
SALT-Trans vs.\ PNAGMDA & deg[100+] F1 & 0.673 & 0.652 & +0.021 & 0.292 & 0.584 \\
SALT-Trans vs.\ PNAGMDA & deg[100+] PR-AUC & 0.683 & 0.672 & +0.011 & 0.471 & 0.584 \\
\bottomrule
\end{tabular}
\end{table*}

\begin{table*}[!htbp]
\centering
\caption{Statistical significance on AMLSim-32k-5\% (node classification).
Runs are paired by split$\times$seed (same train/val/test graph), so we use an exact
paired permutation test (sign-flip, $n=15$ paired runs). We report overall F1 score, the
aggregated high-degree regime (deg[100+]), and the extreme bin (deg[2000--4732]),
against the strongest attention baseline (GAT) and against PNAGMDA (per-layer vs.\
prediction-time fusion). The deg[100+] contrast pools all test predictions with
recipient degree at least 100 and is not the arithmetic mean of the degree-bin F1 scores.
$\Delta$ is the mean paired difference (SALT $-$ baseline).
$p_{\text{Holm}}$ is Holm-corrected within each variant--baseline family.
\protect\sigstar~$p<0.05$, \protect\sigdag~$p<0.10$ (on $p_{\text{Holm}}$).}
\label{tab:significance_amlsim}
\small
\begin{tabular}{@{}llcccc@{}}
\toprule
\textbf{Comparison} & \textbf{Metric} & \textbf{SALT} & \textbf{Base} & \textbf{$\Delta$} & \textbf{$p_{\text{Holm}}$} \\
\midrule
SALT-GAT vs.\ GAT & Overall F1 & 0.855 & 0.821 & +0.034 & $<$0.001\sigstar \\
SALT-GAT vs.\ GAT & Aggregated deg[100+] F1 & 0.617 & 0.479 & +0.138 & $<$0.001\sigstar \\
SALT-GAT vs.\ GAT & deg[2000--4732] F1 & 0.782 & 0.573 & +0.209 & 0.004\sigstar \\
\addlinespace
SALT-GAT vs.\ PNAGMDA & Overall F1 & 0.855 & 0.826 & +0.029 & $<$0.001\sigstar \\
SALT-GAT vs.\ PNAGMDA & Aggregated deg[100+] F1 & 0.617 & 0.512 & +0.105 & $<$0.001\sigstar \\
SALT-GAT vs.\ PNAGMDA & deg[2000--4732] F1 & 0.782 & 0.044 & +0.738 & $<$0.001\sigstar \\
\addlinespace
SALT-Trans vs.\ GAT & Overall F1 & 0.855 & 0.821 & +0.034 & $<$0.001\sigstar \\
SALT-Trans vs.\ GAT & Aggregated deg[100+] F1 & 0.624 & 0.479 & +0.146 & $<$0.001\sigstar \\
SALT-Trans vs.\ GAT & deg[2000--4732] F1 & 0.733 & 0.573 & +0.160 & 0.062\sigdag \\
\addlinespace
SALT-Trans vs.\ PNAGMDA & Overall F1 & 0.855 & 0.826 & +0.029 & $<$0.001\sigstar \\
SALT-Trans vs.\ PNAGMDA & Aggregated deg[100+] F1 & 0.624 & 0.512 & +0.112 & $<$0.001\sigstar \\
SALT-Trans vs.\ PNAGMDA & deg[2000--4732] F1 & 0.733 & 0.044 & +0.689 & $<$0.001\sigstar \\
\bottomrule
\end{tabular}
\end{table*}

Across the three benchmarks, the evidence is strongest on AMLSim, where the paired design yields full statistical power: both SALT variants significantly outperform GAT and PNAGMDA overall and in the high-degree regime ($p_{\text{Holm}}<0.001$ for all but the SALT-Trans extreme-bin contrast, which reaches $p_{\text{Holm}}=0.062$). The contrast with PNAGMDA in the extreme bin is the sharpest: PNAGMDA collapses to $0.044$ while both SALT variants exceed $0.73$ ($\Delta>0.68$, $p_{\text{Holm}}<0.001$), directly supporting per-layer over prediction-time fusion. On HI-Small, SALT-Trans significantly outperforms PNAGMDA on every contrast after correction (overall F1 score, overall PR-AUC, and dense-bin F1 score/PR-AUC; $p_{\text{Holm}}=0.032$ for all four) and FraudGT on overall PR-AUC ($p_{\text{Holm}}=0.032$); On HI-Medium, SALT-Trans significantly outperforms TransConv-84k overall (F1 score and PR-AUC, $p_{\text{Holm}}=0.024$) and in deg[100+] F1 score ($p_{\text{Holm}}=0.043$). Several dense-bin contrasts on the IBM datasets do not reach significance after correction: this is expected, as their high-degree bins contain very few positive examples (see the Fraud Distribution Across Degree Bins section), inflating variance and reducing power. We therefore emphasize effect sizes alongside $p$-values, and note that the consistent direction and magnitude of the gains across all three benchmarks is the central evidence, with formal significance strongest where the data permit it.

\section{Threshold-Free Metric: PR-AUC}
\label{apx:prauc}

Because F1 score depends on the decision threshold (fixed to $0.5$ in the main paper for consistency with prior AML work and across degree bins), we additionally report threshold-free PR-AUC on HI-Small. Table~\ref{tab:smallhi-prauc} shows the same ordering as the F1 score results: SALT-Trans attains the best overall PR-AUC ($0.702$), while SALT-GAT attains the best dense-bin PR-AUC ($0.730$ in deg[50--99]), both improving over FraudGT and the TransConv configurations.
We additionally report degree-stratified PR-AUC on HI-Medium (Table~\ref{tab:fraudgt_mediumhi_prauc_transposed}) and AMLSim-32k-5\% (Table~\ref{tab:amlsim_prauc_full}), computed from the same checkpoints used for the F1 score results. The threshold-free ordering matches the F1 score analysis on both: on HI-Medium, SALT-GAT and SALT-Trans attain the best overall PR-AUC ($0.699$) and lead in the densest populated bin (deg[100+], $0.684$/$0.683$); on AMLSim, both variants are best overall ($0.824$/$0.826$) and dominate the extreme-degree tail (deg[2000--4732], $0.832$/$0.707$). On HI-Medium, where the densest populated bin (deg[100+]) is only at moderate density, PNAGMDA remains competitive in PR-AUC ($0.672$ in deg[100+]); this is consistent with our claim that late fusion degrades specifically in the extreme-density regime, which among our benchmarks is reached only on AMLSim, where PNAGMDA collapses to $0.074$ PR-AUC while SALT exceeds $0.70$.

\begin{table*}[!htbp]
    \centering
    \setlength{\tabcolsep}{4pt}
    \caption{PR-AUC on HI-Small test, stratified by degree (mean $\pm$ std, 5 seeds); \textbf{bold} marks best and \underline{underlined} second best.}
    \label{tab:smallhi-prauc}
    \scriptsize
    \begin{tabular}{@{}lrrrrrr@{}}
    \toprule
    \textbf{Model (Param \#)} & \textbf{deg[1-4]} & \textbf{deg[5-9]} & \textbf{deg[10-19]} & \textbf{deg[20-49]} & \textbf{deg[50-99]} & \textbf{overall} \\
    \midrule
    GIN (73k) & $0.6000 \pm 0.1098$ & $0.3582 \pm 0.0634$ & $0.4892 \pm 0.0753$ & $0.3336 \pm 0.1359$ & $0.5724 \pm 0.1086$ & $0.4104 \pm 0.1051$ \\
    PNA (35k) & $0.6772 \pm 0.0719$ & $0.4396 \pm 0.0292$ & $0.6152 \pm 0.0181$ & $0.5594 \pm 0.0595$ & $0.4946 \pm 0.1021$ & $0.5591 \pm 0.0488$ \\
    TransConv (84k) & $0.7698 \pm 0.0319$ & $0.5062 \pm 0.0169$ & $0.6348 \pm 0.0410$ & $0.6150 \pm 0.0468$ & $0.5848 \pm 0.0760$ & $0.6298 \pm 0.0300$ \\
    TransConv (318k) & $0.7480 \pm 0.0220$ & $0.5110 \pm 0.0140$ & $0.6450 \pm 0.0312$ & $0.6210 \pm 0.0278$ & $0.4903 \pm 0.1004$ & $0.6232 \pm 0.0327$ \\
    FraudGT (182k) & $0.7822 \pm 0.0174$ & $0.5195 \pm 0.0200$ & $\underline{0.6718 \pm 0.0159}$ & $0.6495 \pm 0.0198$ & $0.6588 \pm 0.0590$ & $0.6639 \pm 0.0176$ \\
    PNAGMDA (40k) & $0.8061 \pm 0.0154$ & $0.5189 \pm 0.0249$ & $0.6517 \pm 0.0125$ & $0.6619 \pm 0.0146$ & $0.6596 \pm 0.0284$ & $0.6678 \pm 0.0132$ \\
    \midrule
    SALT-GAT (40k) & $\underline{0.8130 \pm 0.0145}$ & $\underline{0.5407 \pm 0.0235}$ & $\underline{0.6625 \pm 0.0158}$ & $\underline{0.6715 \pm 0.0106}$ & $\textbf{0.7303} \pm \textbf{0.0699}$ & $\underline{0.6853 \pm 0.0086}$ \\
    SALT-Trans (42k) & $\textbf{0.8217} \pm \textbf{0.0205}$ & $\textbf{0.5502} \pm \textbf{0.0152}$ & $\textbf{0.6800} \pm \textbf{0.0144}$ & $\textbf{0.7002} \pm \textbf{0.0141}$ & $\underline{0.7258 \pm 0.0352}$ & $\textbf{0.7019} \pm \textbf{0.0100}$ \\
    \bottomrule
    \end{tabular}
\end{table*}

\begin{table*}[!htbp]
    \centering
    \setlength{\tabcolsep}{4pt}
    \caption{PR-AUC on Medium-HI dataset, stratified by degree (mean $\pm$ std, 3 seeds); \textbf{bold} marks best and \underline{underlined} second best.}
    \label{tab:fraudgt_mediumhi_prauc_transposed}
    \scriptsize
    \begin{tabular}{@{}l *{6}{r} | r@{}}
    \toprule
    \textbf{Model}
    & \textbf{deg[1--4]} & \textbf{deg[5--9]} & \textbf{deg[10--19]} & \textbf{deg[20--49]} & \textbf{deg[50--99]} & \textbf{deg[100+]} & \textbf{Overall} \\
    \midrule
    FraudGT
    & $0.669 \pm 0.149$ & $0.492 \pm 0.028$ & $0.571 \pm 0.103$ & $0.563 \pm 0.061$ & $0.582 \pm 0.070$ & $0.607 \pm 0.183$ & $0.610 \pm 0.088$ \\
    GIN
    & $0.799 \pm 0.029$ & $0.623 \pm 0.026$ & $0.679 \pm 0.019$ & $0.547 \pm 0.024$ & $0.554 \pm 0.033$ & $0.578 \pm 0.042$ & $0.605 \pm 0.026$ \\
    PNA
    & $0.858 \pm 0.006$ & $0.678 \pm 0.035$ & $0.715 \pm 0.030$ & $0.609 \pm 0.026$ & $0.615 \pm 0.054$ & $0.635 \pm 0.056$ & $0.665 \pm 0.029$ \\
    TransConv (84k)
    & $0.840 \pm 0.013$ & $0.674 \pm 0.011$ & $0.714 \pm 0.002$ & $0.604 \pm 0.009$ & $0.625 \pm 0.012$ & $0.637 \pm 0.036$ & $0.661 \pm 0.009$ \\
    TransConv (318k)
    & $0.817 \pm 0.015$ & $0.586 \pm 0.006$ & $0.639 \pm 0.001$ & $0.597 \pm 0.005$ & $0.641 \pm 0.040$ & $0.553 \pm 0.074$ & $0.646 \pm 0.003$ \\
    PNAGMDA
    & $0.841 \pm 0.016$ & $0.692 \pm 0.009$ & $0.712 \pm 0.009$ & $0.604 \pm 0.018$ & $0.631 \pm 0.018$ & $0.672 \pm 0.018$ & $0.664 \pm 0.017$ \\
    \midrule
    SALT-GAT
    & $\textbf{0.879} \pm \textbf{0.013}$ & $\underline{0.724 \pm 0.010}$ & $\underline{0.746 \pm 0.008}$ & $\underline{0.638 \pm 0.021}$ & $\textbf{0.661} \pm \textbf{0.023}$ & $\textbf{0.684} \pm \textbf{0.039}$ & $\textbf{0.699} \pm \textbf{0.018}$ \\
    SALT-Trans
    & $\underline{0.870 \pm 0.012}$ & $\textbf{0.727} \pm \textbf{0.009}$ & $\textbf{0.746} \pm \textbf{0.009}$ & $\textbf{0.640} \pm \textbf{0.008}$ & $\underline{0.660 \pm 0.014}$ & $\underline{0.683 \pm 0.017}$ & $\underline{0.699 \pm 0.009}$ \\
    \bottomrule
    \end{tabular}
\end{table*}

\begin{table*}[hp]
    \centering
    \setlength{\tabcolsep}{2pt}
    \caption{F1 score on AMLSim dataset \citep{GAMLNet}, stratified by degree (mean $\pm$ std, 5 seeds, 3 splits); \textbf{bold} marks best and \underline{underlined} second best. \sigstar{}/\sigdag{} in the extreme-degree bin denote a SALT variant significantly better than the strongest attention baseline (GAT) at $p<0.05$/$p<0.10$ after Holm correction; the full set of significant contrasts (overall and high-degree, vs.\ GAT and PNAGMDA) is reported in the Statistical Significance Tests section of this supplement.}
    \label{tab:amlsim_full}
    \scriptsize
    \begin{tabular}{@{}l *{7}{r} | rr@{}}
    \toprule
    \textbf{Degree Bin}
    & \textbf{GCN} & \textbf{GIN} & \textbf{PNA} & \textbf{GAMLNet} & \textbf{TransConv} & \textbf{GAT} & \textbf{PNAGMDA} & \textbf{SALT-GAT} & \textbf{SALT-Trans} \\
    \midrule
    deg[1--1]
    & $0.986{\pm}0.010$ & $0.977{\pm}0.008$ & $0.989{\pm}0.009$ & $0.977{\pm}0.003$ & $0.952{\pm}0.007$ & $0.976{\pm}0.010$ & $0.988{\pm}0.009$ & $\textbf{0.991}{\pm}\textbf{0.009}$ & $\underline{0.990{\pm}0.007}$ \\
    deg[2--4]
    & $0.709{\pm}0.033$ & $0.656{\pm}0.023$ & $0.783{\pm}0.039$ & $0.745{\pm}0.043$ & $0.743{\pm}0.030$ & $0.740{\pm}0.038$ & $0.706{\pm}0.025$ & $\textbf{0.794}{\pm}\textbf{0.022}$ & $\textbf{0.794}{\pm}\textbf{0.014}$ \\
    deg[5--9]
    & $0.648{\pm}0.002$ & $0.609{\pm}0.008$ & $\textbf{0.841}{\pm}\textbf{0.041}$ & $0.807{\pm}0.023$ & $0.820{\pm}0.047$ & $0.803{\pm}0.042$ & $0.796{\pm}0.038$ & $\textbf{0.842}{\pm}\textbf{0.034}$ & $\underline{0.839{\pm}0.038}$ \\
    deg[10--19]
    & $0.840{\pm}0.022$ & $0.794{\pm}0.022$ & $0.888{\pm}0.020$ & $0.894{\pm}0.020$ & $\underline{0.900{\pm}0.022}$ & $0.886{\pm}0.034$ & $0.851{\pm}0.038$ & $\textbf{0.904}{\pm}\textbf{0.024}$ & $0.893{\pm}0.016$ \\
    deg[20--49]
    & $0.731{\pm}0.058$ & $0.499{\pm}0.069$ & $0.783{\pm}0.020$ & $0.732{\pm}0.050$ & $0.750{\pm}0.049$ & $0.726{\pm}0.013$ & $0.736{\pm}0.030$ & $\underline{0.785{\pm}0.037}$ & $\textbf{0.793}{\pm}\textbf{0.018}$ \\
    deg[50--99]
    & $0.378{\pm}0.075$ & $0.137{\pm}0.118$ & $0.827{\pm}0.034$ & $0.809{\pm}0.018$ & $0.829{\pm}0.028$ & $0.712{\pm}0.079$ & $0.805{\pm}0.053$ & $\underline{0.836{\pm}0.018}$ & $\textbf{0.850}{\pm}\textbf{0.029}$ \\
    deg[100--199]
    & $0.122{\pm}0.062$ & $0.024{\pm}0.007$ & $\underline{0.624{\pm}0.031}$ & $0.538{\pm}0.053$ & $\textbf{0.645}{\pm}\textbf{0.034}$ & $0.478{\pm}0.017$ & $0.592{\pm}0.037$ & $0.617{\pm}0.025$ & $0.639{\pm}0.026$ \\
    deg[200--499]
    & $0.142{\pm}0.058$ & $0.000{\pm}0.000$ & $0.514{\pm}0.097$ & $0.409{\pm}0.046$ & $0.482{\pm}0.100$ & $0.441{\pm}0.118$ & $0.468{\pm}0.073$ & $\textbf{0.532}{\pm}\textbf{0.112}$ & $\underline{0.529{\pm}0.100}$ \\
    deg[500--999]
    & $0.000{\pm}0.000$ & $0.000{\pm}0.000$ & $0.533{\pm}0.189$ & $0.427{\pm}0.050$ & $0.502{\pm}0.166$ & $0.413{\pm}0.207$ & $0.409{\pm}0.354$ & $\textbf{0.556}{\pm}\textbf{0.157}$ & $\textbf{0.556}{\pm}\textbf{0.157}$ \\
    deg[1000--1999]
    & $0.000{\pm}0.000$ & $0.000{\pm}0.000$ & $0.578{\pm}0.166$ & $0.567{\pm}0.233$ & $0.760{\pm}0.200$ & $0.632{\pm}0.260$ & $0.633{\pm}0.289$ & $\textbf{0.827}{\pm}\textbf{0.245}$ & $\underline{0.811{\pm}0.222}$ \\
    deg[2000--4732]
    & $0.000{\pm}0.000$ & $0.000{\pm}0.000$ & $0.178{\pm}0.166$ & $0.000{\pm}0.000$ & $0.442{\pm}0.138$ & $0.573{\pm}0.264$ & $0.044{\pm}0.077$ & $\textbf{0.782}{\pm}\textbf{0.162}$\,\sigstar & $\underline{0.733{\pm}0.196}$\,\sigdag \\
    \midrule
    overall
    & $0.785{\pm}0.011$ & $0.751{\pm}0.012$ & $0.852{\pm}0.008$ & $0.822{\pm}0.010$ & $0.827{\pm}0.008$ & $0.821{\pm}0.008$ & $0.826{\pm}0.005$ & $\textbf{0.855}{\pm}\textbf{0.006}$ & $\underline{0.855{\pm}0.005}$ \\
    \bottomrule
    \end{tabular}
\end{table*}

\begin{table*}[!htbp]
    \centering
    \setlength{\tabcolsep}{2pt}
    \caption{PR-AUC on AMLSim dataset \citep{GAMLNet}, stratified by degree (mean $\pm$ std across split means, 5 seeds, 3 splits); \textbf{bold} marks best and \underline{underlined} second best.}
    \label{tab:amlsim_prauc_full}
    \scriptsize
    \begin{tabular}{@{}l*{9}{r}@{}}
        \toprule
        \textbf{Degree Bin} & \textbf{GCN} & \textbf{GIN} & \textbf{PNA} & \textbf{GAMLNet} & \textbf{TransConv} & \textbf{GAT} & \textbf{PNAGMDA} & \textbf{SALT-GAT} & \textbf{SALT-Trans} \\
        \midrule
        deg[1--1] & $0.996{\pm}0.000$ & $0.997{\pm}0.000$ & $0.998{\pm}0.000$ & $\underline{0.999{\pm}0.000}$ & $\mathbf{0.999}{\pm}\mathbf{0.000}$ & $0.997{\pm}0.000$ & $0.972{\pm}0.000$ & $0.998{\pm}0.000$ & $0.995{\pm}0.000$ \\
        deg[2--4] & $0.649{\pm}0.052$ & $0.629{\pm}0.034$ & $\underline{0.721{\pm}0.067}$ & $0.675{\pm}0.069$ & $0.689{\pm}0.053$ & $0.672{\pm}0.066$ & $0.546{\pm}0.004$ & $0.716{\pm}0.043$ & $\mathbf{0.726}{\pm}\mathbf{0.046}$ \\
        deg[5--9] & $0.625{\pm}0.035$ & $0.629{\pm}0.022$ & $\mathbf{0.839}{\pm}\mathbf{0.058}$ & $0.771{\pm}0.068$ & $0.784{\pm}0.064$ & $0.766{\pm}0.055$ & $0.687{\pm}0.063$ & $0.799{\pm}0.062$ & $\underline{0.828{\pm}0.056}$ \\
        deg[10--19] & $0.839{\pm}0.036$ & $0.836{\pm}0.045$ & $0.885{\pm}0.033$ & $\underline{0.919{\pm}0.050}$ & $\mathbf{0.920}{\pm}\mathbf{0.042}$ & $0.912{\pm}0.055$ & $0.749{\pm}0.047$ & $0.913{\pm}0.050$ & $0.906{\pm}0.047$ \\
        deg[20--49] & $0.699{\pm}0.044$ & $0.567{\pm}0.004$ & $0.776{\pm}0.043$ & $0.756{\pm}0.069$ & $\mathbf{0.797}{\pm}\mathbf{0.050}$ & $0.760{\pm}0.067$ & $0.622{\pm}0.039$ & $0.791{\pm}0.033$ & $\underline{0.793{\pm}0.041}$ \\
        deg[50--99] & $0.305{\pm}0.123$ & $0.197{\pm}0.119$ & $\underline{0.852{\pm}0.010}$ & $0.839{\pm}0.044$ & $0.846{\pm}0.056$ & $0.784{\pm}0.069$ & $0.672{\pm}0.090$ & $\mathbf{0.864}{\pm}\mathbf{0.014}$ & $0.844{\pm}0.049$ \\
        deg[100--199] & $0.201{\pm}0.088$ & $0.127{\pm}0.025$ & $0.631{\pm}0.062$ & $0.567{\pm}0.060$ & $\underline{0.646{\pm}0.047}$ & $0.501{\pm}0.033$ & $0.457{\pm}0.024$ & $0.641{\pm}0.043$ & $\mathbf{0.648}{\pm}\mathbf{0.042}$ \\
        deg[200--499] & $0.297{\pm}0.063$ & $0.108{\pm}0.029$ & $0.494{\pm}0.097$ & $0.496{\pm}0.038$ & $0.477{\pm}0.086$ & $0.469{\pm}0.092$ & $0.391{\pm}0.068$ & $\mathbf{0.547}{\pm}\mathbf{0.116}$ & $\underline{0.526{\pm}0.134}$ \\
        deg[500--999] & $0.204{\pm}0.041$ & $0.143{\pm}0.023$ & $0.576{\pm}0.124$ & $0.459{\pm}0.157$ & $0.535{\pm}0.172$ & $0.521{\pm}0.108$ & $0.414{\pm}0.221$ & $\mathbf{0.610}{\pm}\mathbf{0.101}$ & $\underline{0.608{\pm}0.169}$ \\
        deg[1000--1999] & $0.136{\pm}0.038$ & $0.066{\pm}0.026$ & $0.812{\pm}0.245$ & $0.534{\pm}0.321$ & $0.803{\pm}0.342$ & $0.770{\pm}0.316$ & $0.633{\pm}0.306$ & $\underline{0.819{\pm}0.313}$ & $\mathbf{0.827}{\pm}\mathbf{0.300}$ \\
        deg[2000--4732] & $0.246{\pm}0.213$ & $0.060{\pm}0.017$ & $0.415{\pm}0.223$ & $0.082{\pm}0.033$ & $0.599{\pm}0.223$ & $0.673{\pm}0.300$ & $0.074{\pm}0.036$ & $\mathbf{0.832}{\pm}\mathbf{0.235}$ & $\underline{0.707{\pm}0.254}$ \\
        \midrule
        overall & $0.756{\pm}0.001$ & $0.743{\pm}0.008$ & $0.823{\pm}0.001$ & $0.797{\pm}0.012$ & $0.803{\pm}0.011$ & $0.805{\pm}0.008$ & $0.720{\pm}0.013$ & $\underline{0.824{\pm}0.007}$ & $\mathbf{0.826}{\pm}\mathbf{0.008}$ \\
        \bottomrule
    \end{tabular}
\end{table*}

\begin{table*}[!htbp]
    \centering
    \setlength{\tabcolsep}{5pt}
    \caption{Characteristic-level ablation on HI-Small, full F1 score\,/\,PR-AUC (mean $\pm$ std, 5 seeds). Symbols as in the characteristic-level ablation table of the main paper.}
    \label{tab:ablation_char_full}
    \small
    \begin{tabular}{@{}lcccll@{}}
    \toprule
    \textbf{Model} & \textbf{C1} & \textbf{C2} & \textbf{C3} & \textbf{deg[50--99]} & \textbf{overall} \\
    \midrule
    SALT-Trans (full)        & \checkmark & \checkmark & \checkmark & $0.741\pm0.019 / 0.723\pm0.039$ & $0.716\pm0.018 / 0.700\pm0.014$ \\
    \,\,$\hookrightarrow$ mean+std only & $\circ$ & \checkmark & \checkmark & $0.720\pm0.042 / 0.703\pm0.065$ & $0.712\pm0.004 / 0.674\pm0.013$ \\
    \,\,$\hookrightarrow$ mean only     & $\times$ & \checkmark & \checkmark & $0.636\pm0.065 / 0.613\pm0.079$ & $0.690\pm0.020 / 0.665\pm0.021$ \\
    \,\,$\hookrightarrow$ no degree     & \checkmark & $\times$ & \checkmark & $0.616\pm0.103 / 0.585\pm0.105$ & $0.713\pm0.013 / 0.667\pm0.015$ \\
    \,\,$\hookrightarrow$ no statistics & $\times$ & $\times$ & $\circ$ & $0.637\pm0.032 / 0.589\pm0.031$ & $0.698\pm0.016 / 0.669\pm0.017$ \\
    \midrule
    PNA \textit{(ref.)}      & \checkmark & \checkmark & $\times$ & $0.515\pm0.099 / 0.495\pm0.102$ & $0.629\pm0.032 / 0.559\pm0.049$ \\
    TransConv \textit{(ref.)} & $\times$ & $\times$ & $\circ$ & $0.615\pm0.018 / 0.585\pm0.076$ & $0.645\pm0.031 / 0.629\pm0.030$ \\
    \bottomrule
    \end{tabular}
\end{table*}

\section{Baseline Tuning Robustness}
\label{apx:tuning}

A natural concern is whether the dense-region gap between SALT and the attention baselines is an artifact of insufficient baseline tuning. We address this with three additional studies on HI-Small.

\paragraph{Dense-bin tuning of TransConv and FraudGT.}
We re-tuned TransConv by selecting on densest-bin validation F1 score rather than overall validation F1 score. The best configuration reached $0.708$ on deg[50--99] validation (lr$=0.005$, hidden$=64$, class weight $=6.156$, dropout$=0.131$), but only $0.570\pm0.024$ on the deg[50--99] test set and $0.604\pm0.017$ overall, still well below SALT-Trans. FraudGT showed the same trend. On AMLSim, tuning over the dense tail (the last bin had too few positives to tune on directly) yielded F1 score of $0.638/0.503/0.514/0.771/0.533$ from deg[100--199] to deg[2000--4732]: mixed and, on aggregate, below the SALT variants evaluated against the globally tuned TransConv. Tuning the baselines specifically for dense bins therefore does not close the gap.

\paragraph{Attention-temperature scaling.}
We also tested whether manually forcing TransConv to spread its attention wider (via softmax temperature scaling) resolves its dense-bin failure. It does not, and broader attention does not monotonically improve dense-bin F1 score: at temperature $4$, TransConv reached $0.443\pm0.098$ on deg[50--99] and $0.599\pm0.008$ overall; at temperature $2$, $0.455\pm0.214$ and $0.673\pm0.012$. This supports our reading of the \ENNs{} diagnostics: the problem is not the raw spread of attention but whether weak fraud-relevant evidence is preserved.

\paragraph{Parameter-matched ablation.}
Finally, to rule out model size as the explanation, we evaluated a TransConv reduced to $\approx$40k parameters (matching SALT). It reached overall F1 score $0.624\pm0.017$ and $0.544\pm0.048$ on deg[50--99], confirming that SALT's gains are not explained by parameter count alone.

\section{Final Residual-Step Ablation}
\label{apx:residual}

The final update averages the previous state and the fusion output, $\mathbf{h}_v^{(k)}=(\mathbf{h}_v^{(k-1)}+\mathbf{h}_{v,\mathrm{fusion}}^{(k)})/2$. We keep this residual integration consistent with other AML GNNs so that gains cannot be attributed to removing the usual residual step. Removing the final residual (using $\mathbf{h}_v^{(k)}=\mathbf{h}_{v,\mathrm{fusion}}^{(k)}$ directly) on HI-Small slightly destabilizes training and reduces dense-bin F1 score, while the per-layer fusion remains the dominant source of the dense-region improvement; the previous-state term in the concatenation $\mathbf{z}_v^{(k)}$ is what supplies the fusion MLP with context, not the residual averaging alone.

\section{Hyperparameter Configuration}
\label{apx:hyperparameters}
All models are trained with the Adam optimizer. Unless otherwise stated, the optimization objective is the F1 score of the minority (fraudulent) class on the validation set. Table~\ref{tab:hyper_provenance} summarizes, for each model and dataset, whether the configuration was \emph{reused} from prior work, \emph{tuned by us}, or \emph{selected via full grid search}.
\begin{table*}[!htbp]
    \centering
    \caption{Provenance of hyperparameter configurations. Reused: published tuned configs adopted as-is; Tuned: tuned by us over the ranges below; Grid: full grid search with the ranges below.}
    \label{tab:hyper_provenance}
    \scriptsize
    \setlength{\tabcolsep}{4pt}
    \begin{tabular}{@{}l p{0.40\textwidth} p{0.42\textwidth}@{}}
    \toprule
    \textbf{Model} & \textbf{HI-Small / HI-Medium} & \textbf{AMLSim-32k-5\%} \\
    \midrule
    GIN       & Grid \citep{IBMdatasets}        & Grid \citep{GAMLNet} \\
    GCN       & --- (no edge features)            & Grid \citep{GAMLNet}\\
    PNA       & Grid \citep{IBMdatasets}        & Grid \citep{GAMLNet} \\
    GAT       & Grid \citep{IBMdatasets}        & Grid \citep{GAMLNet} \\
    TransConv & Tuned with similar grid of \citep{IBMdatasets} (lr; hidden 64/128)         & Grid \\
    FraudGT   & Grid by authors \citep{fraudGT} with similar grid of \citep{IBMdatasets}           & --- \\
    GAMLNet   & ---                               & Grid \citep{GAMLNet} \\
    PNAGMDA   & Inherits PNA/GAT backbone       & Inherits PNA/GAT backbone (only hidden size adapted for matching heads and towers) \\
    SALT-GAT  & Inherits PNA/GAT backbone         & Inherits PNA/GAT backbone (only hidden size adapted for matching heads and towers) \\
    SALT-Trans& Inherits PNA/TransConv backbone   & Inherits PNA/TransConv backbone (only hidden size adapted for matching heads and towers) \\
    \bottomrule
    \end{tabular}
\end{table*}

\subsection{HI-Small and HI-Medium}
\label{apx:hyperparameters_hi}

Both HI-Small and HI-Medium adopt a \emph{strictly temporal split}: transactions are ordered by timestamp and partitioned into non-overlapping training, validation, and test intervals. Consequently, the data distribution (including fraud distribution across degree bins) is fixed and identical across runs and seeds.

For these benchmarks, we followed standard practice in prior AML work and reused the published baseline configurations consistently adopted across these datasets~\citep{IBMdatasets, fraudGT, multiGNNproofs, multiGraphAggr}, rather than introducing a new full re-tuning pipeline for already-established baselines. A full grid search at this scale is impractical: a conservative grid (4 learning rates, 2 depths, 3 hidden sizes, 3 dropouts, 3 class weights) would require $>$2000 GPU-hours on HI-Small and $>$13000 on HI-Medium on a single RTX A6000, which is precisely why prior AML-GNN work reuses published tuned configs. The reference search ranges are:

\begin{center}
\begin{tabular}{@{}ll@{}}
\toprule
\textbf{Hyperparameter} & \textbf{Search range} \\
\midrule
hidden dim.   & 16--72 \\
learning rate & 0.0001--0.05 \\
\# GNN layers & 2--4 \\
dropout       & 0.0--0.5 \\
class weight  & 6--8 \\
\bottomrule
\end{tabular}
\end{center}

\noindent\textbf{Models evaluated.}
\begin{itemize}
    \item PNA, GIN, GAT
    \item TransConv (84k parameters, hidden size $64$)
    \item TransConv (318k parameters, hidden size $128$)
    \item FraudGT, PNAGMDA
    \item SALT-GAT, SALT-Trans
\end{itemize}

\noindent\textbf{TransConv configurations.}
Since no canonical tuned configuration exists for these IBM benchmarks, we tuned TransConv ourselves. We fixed the depth to 2 layers, in line with all baselines on these datasets, to test whether the degradation is mainly a width/capacity issue rather than a full model-selection effect. TransConv inherits the same architectural hyperparameters as the GAT tuned by \citet{IBMdatasets} (number of heads and dropout), which are consistent with those found by \citet{fraudGT} for FraudGT, differing only in the attention operator and learning rate. After a coarse search, we fine-tuned the learning rate $\in \{0.006, 0.005, 0.003, 0.002, 0.001, 0.0005\}$ for hidden sizes 64 and 128:
\begin{itemize}
    \item \textbf{84k parameters}: hidden dimension $64$;
    \item \textbf{318k parameters}: hidden dimension $128$.
\end{itemize}

\noindent\textbf{SALT variants and PNAGMDA.}
SALT-GAT inherits all PNA hyperparameters (depth, hidden size, aggregation, normalization, loss weighting). SALT-Trans is identical to PNA in all hyperparameters, with the learning rate fixed to $0.001$. PNAGMDA reuses the SALT-GAT backbone hyperparameters. SALT performs no model-specific tuning, so performance differences are attributable to architectural design rather than additional tuning advantages.

\noindent\textbf{Final HI-Small / HI-Medium configurations.}
All models use the Adam optimizer, cross-entropy loss with positive-class
weight $w$, z-normalization, $2$ GNN layers, and a single MLP layer; both
datasets share the same configuration per model. The two TransConv variants
differ only in hidden size ($64$ vs.\ $128$), as noted above.
\begin{itemize}
    \item \textbf{GIN:} lr$=0.0062$, $F=64$, $w=6$, dropout$=0.010$, final dropout$=0.105$.
    \item \textbf{PNA:} lr$=0.0006$, $F=20$, $w=7$, dropout$=0.083$, final dropout$=0.288$.
    \item \textbf{GAT:} lr$=0.006$, $F=64$, $H=4$, $w=6$, dropout$=0.009$, final dropout$=0.1$.
    \item \textbf{TransConv (84k):} lr$=0.001$, $F=64$, $H=4$, $w=6$, dropout$=0.009$, final dropout$=0.1$.
    \item \textbf{TransConv (318k):} as above, with $F=128$.
    \item \textbf{FraudGT:} lr$=0.001$ (AdamW, cosine schedule with $5$ warmup epochs, weight decay $10^{-5}$), $F=64$, $H=8$, $w=6$, $2$ transformer layers ($+2$ post-GT layers), transformer dropout$=0.2$, attention dropout$=0.3$; edge-masked sparse-node attention with layer norm and GELU.
    \item \textbf{SALT-GAT:} lr$=0.0006$, $F=20$, $w=7$, dropout$=0.083$, final dropout$=0.288$.
    \item \textbf{SALT-Trans:} lr$=0.0012$, $F=20$, $w=7$, dropout$=0.083$, final dropout$=0.288$.
    \item \textbf{PNAGMDA:} lr$=0.0006$, $F=20$, $w=7$, dropout$=0.083$, final dropout$=0.288$.
\end{itemize}
\subsection{AMLSim dataset}
\label{apx:hyperparameters_amlsim}

\noindent\textbf{Hyperparameter search.}
For models not already optimized in prior work, we perform a grid search over:
\begin{itemize}
    \item learning rate $\in \{0.01, 0.005, 0.001\}$;
    \item number of layers $K \in \{2,3,4,5,6\}$;
    \item hidden dimension $F \in \{8,16,32,64\}$;
    \item class-weight parameter $w \in \{0.26, 0.44, 0.79, 0.96\}$;
    \item heads $\in \{2, 4, 8\}$ when applicable (GAT/TransConv);
    \item attention dropout $\in \{0.0, 0.1, 0.2, 0.3\}$ when applicable.
\end{itemize}
Each configuration is evaluated over multiple seeds; the best model is selected by validation F1 score.

\noindent\textbf{Final AMLSim configurations.}
\begin{itemize}
    \item \textbf{GAMLNet:} lr$=0.01$, $w=0.44$, $K_1=2,\ K_2=6$, $F_1=8,\ F_2=64$ (GIN and GraphSAGE components).
    \item \textbf{GIN:} lr$=0.01$, $w=0.44$, $K=1$, $F=8$.
    \item \textbf{GCN:} lr$=0.01$, $w=0.44$, $K=3$, $F=32$.
    \item \textbf{GAT:} lr$=0.01$, $w=0.79$, $K=2$, $F=64$, $H=2$, dropout$=0.0$.
    \item \textbf{TransConv:} lr$=0.005$, $w=0.44$, $K=3$, $F=64$, $H=2$, dropout$=0.3$.
    \item \textbf{PNA:} lr$=0.005$, $w=0.96$, $K=2$, $F=20$.
    \item \textbf{SALT-GAT:} lr$=0.005$, $w=0.79$, $K=2$, $F=40$, dropout$=0.0$.
    \item \textbf{SALT-Trans:} lr$=0.005$, $w=0.44$, $K=2$, $F=40$, dropout$=0.3$.
\end{itemize}

Prior works on AMLSim adopt a \emph{random split}. Random partitioning can yield degenerate splits in which some degree bins (especially dense ones) contain no fraudulent examples in train/val/test, biasing the evaluation.  We apply a pre-defined validity filter to avoid degenerate random splits where the target degree-stratified metric is undefined:
\begin{itemize}
    \item all degree bins, including the densest, contain fraudulent examples in train, validation, and test;
    \item each split contains, in each bin, at least one recognizable fraud pattern, i.e., patterns provably detectable by vanilla GNNs without reverse message passing, ports, or ego identifiers~\citep{multiGNNproofs}, such as cycles, fan-in, scatter-gather, and gather-scatter.
\end{itemize}
All AMLSim results are averaged over these three splits.

\section{Fraud Distribution Across Degree Bins}
\label{apx:fraud_distribution}

For HI-Small and HI-Medium, the split is \emph{strictly temporal}; the fraud distribution is therefore fixed and identical across runs and seeds. Tables~\ref{tab:others_only_fraud}, \ref{tab:in_degree_stats_complete}, and~\ref{tab:in_degree_stats_mediumHI_complete} report the fraud distribution across in-degree bins. Both datasets concentrate fraudulent activity in medium-to-high degree bins, underscoring the importance of handling dense neighborhoods where legitimate and illicit transactions are tightly interwoven.

\begin{table*}[!htbp]
\centering
\caption{Agent-based fraud distribution across degree bins for HI-Small.}
\label{tab:others_only_fraud}
\small
\begin{tabular}{@{}lrrrrr@{}}
\toprule
\textbf{Split} &
\begin{tabular}[c]{@{}c@{}}\textbf{Degree}\\\textbf{Bin}\end{tabular} &
\begin{tabular}[c]{@{}c@{}}\textbf{Fraud}\\\textbf{Count}\end{tabular} &
\begin{tabular}[c]{@{}c@{}}\textbf{Fraud}\\\textbf{Share \%}\end{tabular} &
\begin{tabular}[c]{@{}c@{}}\textbf{Total}\\\textbf{Count}\end{tabular} &
\begin{tabular}[c]{@{}c@{}}\textbf{Fraud}\\\textbf{Rate \%}\end{tabular} \\
\midrule
\multirow{5}{*}{\textit{Train}} & [1,\phantom{1}4]  & 186 & 15.30 & 390,746   & 0.05 \\
                               & [5,\phantom{1}9]  & 315 & 25.90 & 254,855   & 0.12 \\
                               & [10,19]           & 480 & 39.47 & 1,616,927 & 0.03 \\
                               & [20,49]           & 223 & 18.34 & 971,372   & 0.02 \\
                               & [50,99]           & 12  & 0.99  & 12,834    & 0.09 \\
\midrule
\multirow{5}{*}{\textit{Test}}  & [1,\phantom{1}4]  & 53  & 14.29 & 40,576    & 0.13 \\
                               & [5,\phantom{1}9]  & 80  & 21.56 & 41,211    & 0.19 \\
                               & [10,19]           & 86  & 23.18 & 136,546   & 0.06 \\
                               & [20,49]           & 139 & 37.47 & 589,963   & 0.02 \\
                               & [50,99]           & 13  & 3.50  & 53,035    & 0.02 \\
\bottomrule
\end{tabular}
\end{table*}

\begin{table*}[!htbp]
\centering
\caption{Complete in-degree binning analysis across all splits on HI-Small.}
\label{tab:in_degree_stats_complete}
\scriptsize
\setlength{\tabcolsep}{4pt}
\begin{tabular}{@{}l r r r r r@{}}
\toprule
\textbf{Split (scope)} & \textbf{Bin's in-degree} & \textbf{Fraud \#} & \textbf{Fraud share \%} & \textbf{Total \#} & \textbf{Fraud rate \%} \\
\midrule
\multirow{6}{*}{train (train)}
  & [1,\phantom{1}4]     & 503 & 19.88 & 391{,}063  & 0.13 \\
  & [5,\phantom{1}9]     & 459 & 18.14 & 254{,}999  & 0.18 \\
  & [10,19]             & 892 & 35.26 & 1{,}617{,}339 & 0.06 \\
  & [20,49]             & 664 & 26.25 & 971{,}813  & 0.07 \\
  & [50,99]             &  12 &  0.47 & 12{,}834   & 0.09 \\
  & [100+]              &   0 &  0.00 &    873     & 0.00 \\
\midrule
\multirow{6}{*}{val (train+val)}
  & [1,\phantom{1}4]     & 232 & 22.39 &   5{,}676  & 4.09 \\
  & [5,\phantom{1}9]     & 132 & 12.74 &  23{,}460  & 0.56 \\
  & [10,19]             & 275 & 26.54 & 319{,}887  & 0.09 \\
  & [20,49]             & 386 & 37.26 & 596{,}971  & 0.06 \\
  & [50,99]             &  10 &  0.97 &  19{,}356  & 0.05 \\
  & [100+]              &   1 &  0.10 &     174    & 0.57 \\
\midrule
\multirow{6}{*}{test (full)}
  & [1,\phantom{1}4]     & 314 & 19.49 &  40{,}837  & 0.77 \\
  & [5,\phantom{1}9]     & 204 & 12.66 &  41{,}335  & 0.49 \\
  & [10,19]             & 333 & 20.67 & 136{,}793  & 0.24 \\
  & [20,49]             & 636 & 39.48 & 590{,}460  & 0.11 \\
  & [50,99]             & 124 &  7.70 &  53{,}146  & 0.23 \\
  & [100+]              &   0 &  0.00 &   1{,}329  & 0.00 \\
\bottomrule
\end{tabular}
\end{table*}

\begin{table*}[!htbp]
\centering
\caption{Complete in-degree binning analysis across all splits on HI-Medium.}
\label{tab:in_degree_stats_mediumHI_complete}
\scriptsize
\setlength{\tabcolsep}{4pt}
\begin{tabular}{@{}l r r r r r@{}}
\toprule
\textbf{Split (scope)} & \textbf{Bin's in-degree} & \textbf{Fraud \#} & \textbf{Fraud share \%} & \textbf{Total \#} & \textbf{Fraud rate \%} \\
\midrule
\multirow{6}{*}{train (train)}
  & [1,\phantom{1}4]     & 2{,}944 & 18.74 & 1{,}535{,}071  & 0.19 \\
  & [5,\phantom{1}9]     & 2{,}426 & 15.44 &   727{,}528    & 0.33 \\
  & [10,19]             & 3{,}186 & 20.28 & 3{,}789{,}170  & 0.08 \\
  & [20,49]             & 6{,}671 & 42.46 & 12{,}584{,}957 & 0.05 \\
  & [50,99]             &   478   &  3.04 &   836{,}730    & 0.06 \\
  & [100+]              &     8   &  0.05 &    13{,}670    & 0.06 \\
\midrule
\multirow{6}{*}{val (train+val)}
  & [1,\phantom{1}4]     & 1{,}566 & 18.42 &    36{,}478    & 4.29 \\
  & [5,\phantom{1}9]     & 1{,}304 & 15.34 &    53{,}375    & 2.44 \\
  & [10,19]             & 1{,}156 & 13.60 &   388{,}330    & 0.30 \\
  & [20,49]             & 3{,}673 & 43.21 & 3{,}978{,}480  & 0.09 \\
  & [50,99]             &   780   &  9.18 & 1{,}054{,}339  & 0.07 \\
  & [100+]              &    22   &  0.26 &     9{,}881    & 0.22 \\
\midrule
\multirow{6}{*}{test (full)}
  & [1,\phantom{1}4]     & 1{,}780 & 16.16 &   109{,}099    & 1.63 \\
  & [5,\phantom{1}9]     & 1{,}418 & 12.87 &   226{,}761    & 0.63 \\
  & [10,19]             & 1{,}398 & 12.69 &   298{,}687    & 0.47 \\
  & [20,49]             & 4{,}085 & 37.08 & 3{,}707{,}509  & 0.11 \\
  & [50,99]             & 2{,}263 & 20.54 & 2{,}455{,}117  & 0.09 \\
  & [100+]              &    72   &  0.65 &    93{,}056    & 0.08 \\
\bottomrule
\end{tabular}
\end{table*}

\section{Time Performance}
\label{apx:time}

\begin{figure}[!htbp]
  \centering
  \includegraphics[width=.8\linewidth]{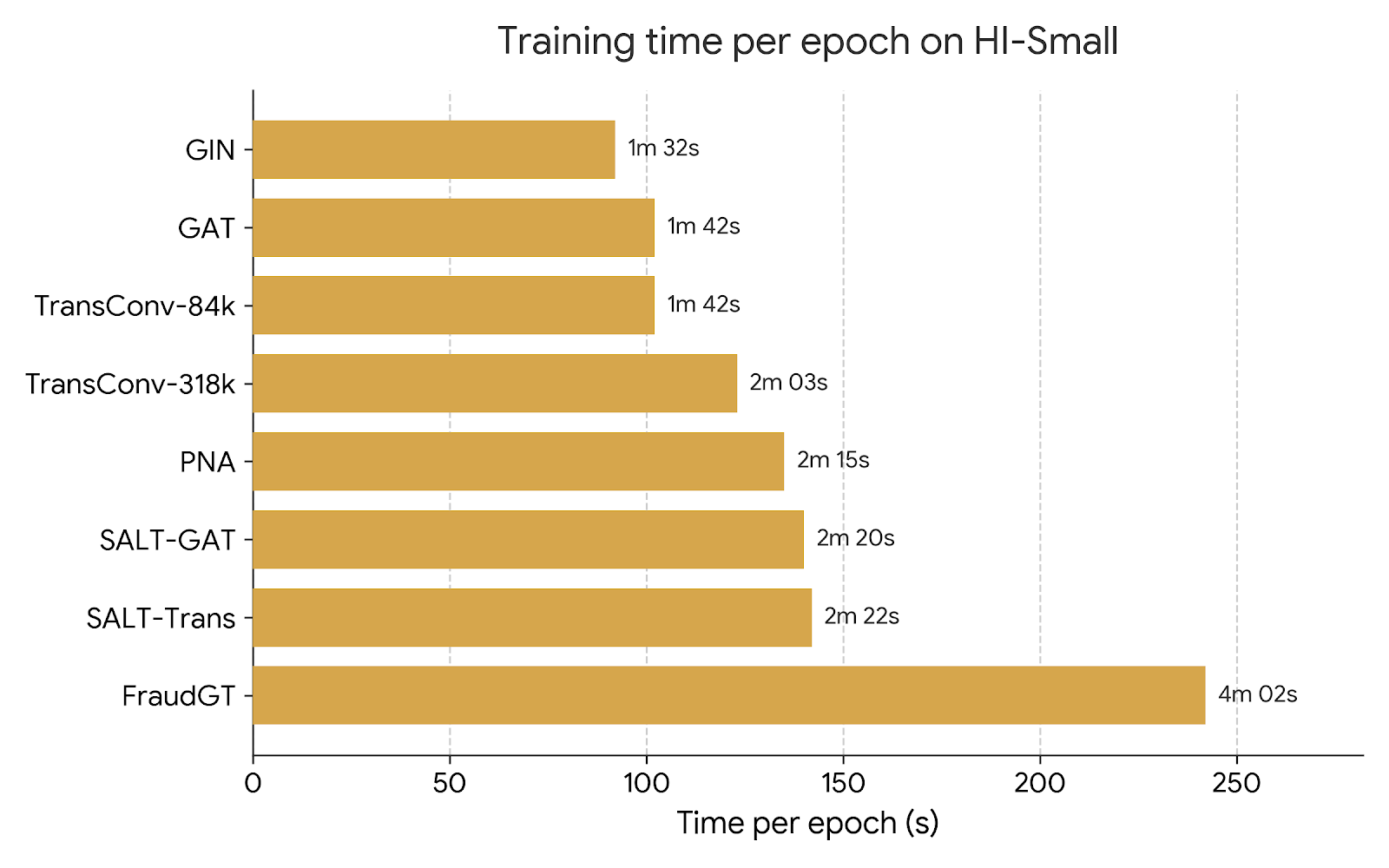}
  \caption{Average training time per epoch for HI-Small on a single NVIDIA RTX A6000 with batch size $8{,}192$ and 100 neighbors per hop.}
  \label{fig:time_epoch}
\end{figure}

Despite its hybrid design, SALT exhibits a training cost comparable to PNA and substantially lower than high-capacity TransformerConv models, confirming that improved detection performance does not come at the expense of prohibitive computational overhead.

\section{Algorithms}
\label{apx:algorithms}

For completeness, we provide pseudocode for the two procedures referenced in the main paper. Algorithm~\ref{alg:degstrat} details the recipient-degree stratified evaluation protocol, and Algorithm~\ref{alg:salt} details the layer-wise SALT-GNN forward pass.

\begin{algorithm}[!htbp]
\small
\caption{Degree-Stratified Evaluation}
\label{alg:degstrat}
\begin{algorithmic}[1]
\REQUIRE Graph $\mathcal{G}=(\mathcal{V},\mathcal{E})$; trained model $f$; bin edges $\mathcal{B}=\{b_0,b_1,\ldots\}$ (log-spaced); task $\tau\in\{\text{node},\text{edge}\}$
\ENSURE Per-bin scores $\{F1_b\}$; (attention models) $\{\mathrm{ENNs}_b^{(\ell)}, \mathrm{Hit@}k_b\}$
\STATE Initialize empty prediction set $P_b\leftarrow\emptyset$ for each bin $b$
\FORALL{target $t$ (node $v$ or transaction $(u,v)$) in the test set}
    \STATE $v \leftarrow t$ if $\tau=\text{node}$ else receiver of $t$ \COMMENT{receiver aggregates incoming evidence}
    \STATE $d_v \leftarrow |\mathcal{N}_{\text{in}}(v)|$
    \STATE $b \leftarrow \textsc{Bin}(d_v,\mathcal{B})$
    \STATE $P_b \leftarrow P_b \cup \{(f(t), y_t)\}$ \COMMENT{store (prediction, label)}
\ENDFOR
\FORALL{bin $b$}
    \STATE $F1_b \leftarrow \textsc{F1}(P_b)$ at threshold $0.5$
\ENDFOR
\IF{$f$ uses attention}
    \FORALL{node $i$, layer $\ell$}
        \STATE $\boldsymbol{\alpha}_i^{(\ell)} \leftarrow$ attention weights over $\mathcal{N}(i)$
        \STATE $H \leftarrow -\sum_{j}\alpha_{ij}^{(\ell)}\log\alpha_{ij}^{(\ell)}$
        \STATE $\mathrm{ENNs}_i^{(\ell)} \leftarrow \exp(H)$ \COMMENT{$=d_i$ if uniform, $1$ if peaked}
    \ENDFOR
    \FORALL{node $i$ with $\geq 1$ fraudulent in-neighbor}
        \STATE $\mathrm{hit}_i \leftarrow \mathbf{1}[\exists\text{ fraud edge in top-}k\text{ of }\boldsymbol{\alpha}_i]$
    \ENDFOR
    \STATE Aggregate $\mathrm{ENNs}_b^{(\ell)}$, $\mathrm{Hit@}k_b$ per bin $b$
\ENDIF
\RETURN $\{F1_b\}$ and attention diagnostics
\end{algorithmic}
\end{algorithm}

\begin{algorithm}[!htbp]
\small
\caption{SALT-GNN Layer-wise Forward Pass}
\label{alg:salt}
\begin{algorithmic}[1]
\REQUIRE Graph $\mathcal{G}$; edge features $\mathbf{e}$; input features $\mathbf{h}_v^{(0)}$; layers $K$; aggregators $\mathcal{A}=\{\mu,\sigma,\min,\max\}$; degree scalers $\mathcal{S}$
\ENSURE Final node embeddings $\mathbf{h}_v^{(K)}$
\FOR{$k=1$ \TO $K$}
    \FORALL{node $v\in\mathcal{V}$}
        \STATE \textit{// Statistical channel (Characteristics 1--2)}
        \STATE $\mathbf{a}_v \leftarrow \bigoplus_{s\in\mathcal{S}}\bigoplus_{\mathrm{agg}\in\mathcal{A}} s(d_v)\cdot\mathrm{agg}\big(\{\mathbf{h}_u^{(k-1)}\!:u\in\mathcal{N}_v\}\big)$
        \STATE $\mathbf{h}_{v,\mathrm{stat}}^{(k)} \leftarrow \mathrm{U}_{\mathrm{stat}}^{(k)}\big(\mathbf{h}_v^{(k-1)}\,\Vert\,\mathbf{a}_v\big)$
        \STATE \textit{// Attention channel (selectivity)}
        \STATE $\alpha_{v,u}^{(k)} \leftarrow \mathrm{softmax}_{u\in\mathcal{N}_v}\big(\text{score}(\mathbf{h}_v^{(k-1)},\mathbf{h}_u^{(k-1)},\mathbf{e}_{v,u})\big)$
        \STATE $\mathbf{h}_{v,\mathrm{att}}^{(k)} \leftarrow \sum_{u\in\mathcal{N}_v}\alpha_{v,u}^{(k)}\big(\mathbf{h}_u^{(k-1)}+\mathbf{e}_{v,u}\big)$
        \STATE \textit{// Layer-wise fusion (interaction during propagation)}
        \STATE $\mathbf{z}_v^{(k)} \leftarrow \big[\mathbf{h}_{v,\mathrm{stat}}^{(k)}\,\Vert\,\mathbf{h}_{v,\mathrm{att}}^{(k)}\,\Vert\,\mathbf{h}_v^{(k-1)}\big]$
        \STATE $\mathbf{h}_{v,\mathrm{fusion}}^{(k)} \leftarrow \mathrm{MLP}_{\mathrm{fusion}}^{(k)}\big(\mathbf{z}_v^{(k)}\big)$
        \STATE $\mathbf{h}_v^{(k)} \leftarrow \big(\mathbf{h}_v^{(k-1)}+\mathbf{h}_{v,\mathrm{fusion}}^{(k)}\big)/2$
    \ENDFOR
\ENDFOR
\RETURN $\{\mathbf{h}_v^{(K)}\}$
\end{algorithmic}
\end{algorithm}

\section{Controlled Synthetic Validation}
\label{sec:synthetic}

The degree-stratified results in the main paper are observational: they
localize \emph{where} GNNs degrade but cannot, on real data alone, attribute the
degradation to a specific aggregation property. To complement them, we build a
small controlled testbed of synthetic transaction graphs with \emph{planted ground
truth}, in which each of the three characteristics analyzed in the main paper can be
isolated and turned on or off. All models are instantiated from a single backbone
that differs only in the component under test (statistical channel, attention
operator, degree scalers, and per-layer vs.\ prediction-time fusion); every task
includes a feature-only floor (a classifier with no graph access, expected at
chance) and an oracle that scores the planted rule (the attainable ceiling). The
generators, models, and protocol are released with our code.

\paragraph{Characteristics~1--2 (multiset and cardinality).}
We construct two families of graphs. In the \emph{multiset} family, positive and
negative accounts are built to share the same neighborhood mean while differing in
the rest of the multiset, so that a single mean summary collapses them; in the
\emph{cardinality} family, all incoming transactions are made feature-identical and
only their \emph{count} differs, so that a normalized aggregator cannot recover the
signal. Table~\ref{tab:synth-c12} reports the results. A single mean aggregator and
the feature-only floor collapse to chance on both, whereas multi-aggregation
recovers the multiset signal and degree-aware scaling recovers the cardinality
signal. The comparison is causal at the component level: removing only the degree
scalers from SALT (``SALT-Trans $-$ degree'') reproduces the cardinality failure,
while the full model matches the oracle. Attention alone passes the multiset case
(it can emphasize extreme neighbors) but, like the mean aggregator, remains blind to
cardinality, mirroring the behavior of the attention baselines on dense real
neighborhoods.

\begin{table*}[!htbp]
\centering
\caption{Controlled synthetic validation of Characteristics~1--2. Candidate-edge
PR-AUC ($\uparrow$; $0.5$ = chance, oracle $=1.0$), mean over seeds. \textbf{Bold}
marks a model collapsing to chance on that axis. ``$-$~degree'' removes only the
degree scalers from SALT (a component-level knockout).}
\label{tab:synth-c12}
\begin{tabular}{@{}lcc@{}}
\toprule
\textbf{Model} & \textbf{Multiset (C1)} & \textbf{Cardinality (C2)} \\
\midrule
Feature-only floor (no graph)      & \textbf{0.50} & \textbf{0.50} \\
Single mean aggregator             & \textbf{0.54} & \textbf{0.51} \\
Attention-only (TransConv)         & 1.00          & \textbf{0.53} \\
SALT-Trans $-$ degree (knockout)   & 1.00          & \textbf{0.57} \\
PNA                                & 1.00          & 0.99 \\
SALT-Trans                         & 1.00          & 0.99 \\
\midrule
Oracle (planted rule)              & 1.00          & 1.00 \\
\bottomrule
\end{tabular}
\end{table*}

\paragraph{Characteristic~3 (layer-wise fusion and attention routing).}
The third characteristic concerns whether weak but pattern-relevant evidence remains
available to attention across hops. A pure task-accuracy separation is not the right
test here: a PNA-containing late-fusion model can often solve an existence-style
synthetic label through its statistical branch, even if its attention branch never
routes mass to the planted fraud. We therefore measure the mechanism directly.

\begin{figure}[!htbp]
\centering
\includegraphics[width=\linewidth]{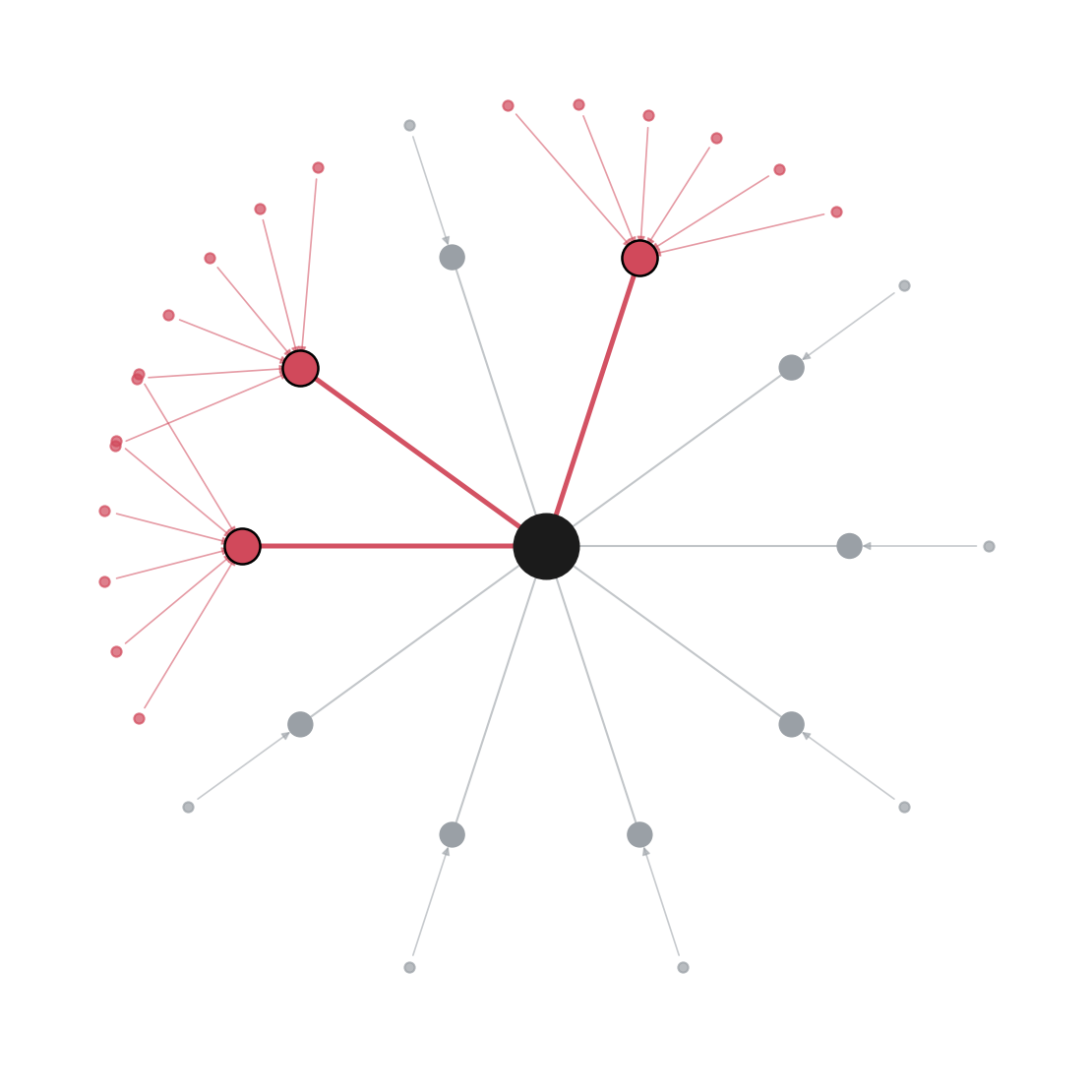}
\caption{A constructed two-hop fraud motif. A receiver (hub) collects from many
mules; a few \emph{coordinated} mules (red) are the planted fraud, distinguishable
only by a first-hop statistic of their own upstream (here, high cardinality), while
legitimate mules (gray) have a sparse upstream. The fraud is rare among the
neighbors, as on real high-degree accounts.}
\label{fig:synth-motif}
\end{figure}

We plant a two-hop motif (Fig.~\ref{fig:synth-motif}) in which a dense receiver
collects from $D=40$ mules. Positive receivers hide one to three coordinated mules;
negative receivers hide none. A mule is coordinated only if its own upstream
neighborhood has a high-cardinality signature, so the cue is unavailable to the
receiver's attention at the first hop. The graded probe edge asks whether the
receiver contains any coordinated mule. We report probe PR-AUC only as a sanity
check. The C3 metric is the attention AUC over incoming mule edges on positive test
receivers: $0.5$ is random ranking and $1.0$ means the coordinated mule is ranked
above all legitimate mules. We also report Hit@10, matching the main-paper
diagnostic.

Table~\ref{tab:synth-c3} shows the result. At L0 all attention models are random,
because coordinated and legitimate mules have not yet incorporated their upstream
statistics. At L1, only per-layer SALT routes attention to the planted fraud. The
strict causal control is \emph{SALT-Trans $-$ per-layer}, which keeps the same
statistical and Transformer-attention operators but moves fusion to prediction time;
it still solves the probe label (PR-AUC $0.995$) yet its L1 attention remains random.
PNAGMDA, the external late-fusion precedent, also solves the probe label but does not
reliably route attention to the coordinated mules. Thus the experiment does not
claim a representational impossibility for late fusion; it isolates the mechanism
SALT needs in dense AML graphs: statistical evidence must enter the state before the
next attention step, not only at the final classifier.

\begin{table*}[!htbp]
\centering
\caption{Controlled C3 routing test on the planted two-hop motif ($D=40$, 5 seeds).
Probe PR-AUC confirms that late-fusion statistical branches can solve the label;
attention AUC and Hit@10 measure whether the receiver's attention ranks the planted
coordinated mule highly. Only layer-wise SALT moves the L1 attention from random
ranking to perfect routing.}
\label{tab:synth-c3}
\small
\begin{tabular}{@{}lcccc@{}}
\toprule
\textbf{Model} & \textbf{Fusion} & \textbf{Probe PR-AUC} & \textbf{Attn AUC L0/L1} & \textbf{Hit@10 L1} \\
\midrule
TransConv                  & attention only  & $0.556\pm0.078$ & $0.500\pm0.000\ /\ 0.500\pm0.000$ & $0.400\pm0.071$ \\
PNAGMDA                    & prediction-time & $1.000\pm0.000$ & $0.500\pm0.000\ /\ 0.205\pm0.398$ & $0.200\pm0.400$ \\
SALT-Trans $-$ per-layer   & prediction-time & $0.995\pm0.010$ & $0.500\pm0.000\ /\ 0.500\pm0.000$ & $0.400\pm0.071$ \\
SALT-Trans                 & layer-wise      & $\textbf{1.000}\pm\textbf{0.000}$ & $0.500\pm0.000\ /\ \textbf{1.000}\pm\textbf{0.000}$ & $\textbf{1.000}\pm\textbf{0.000}$ \\
\bottomrule
\end{tabular}
\end{table*}

\begin{figure}[!htbp]
\centering
\includegraphics[width=\linewidth]{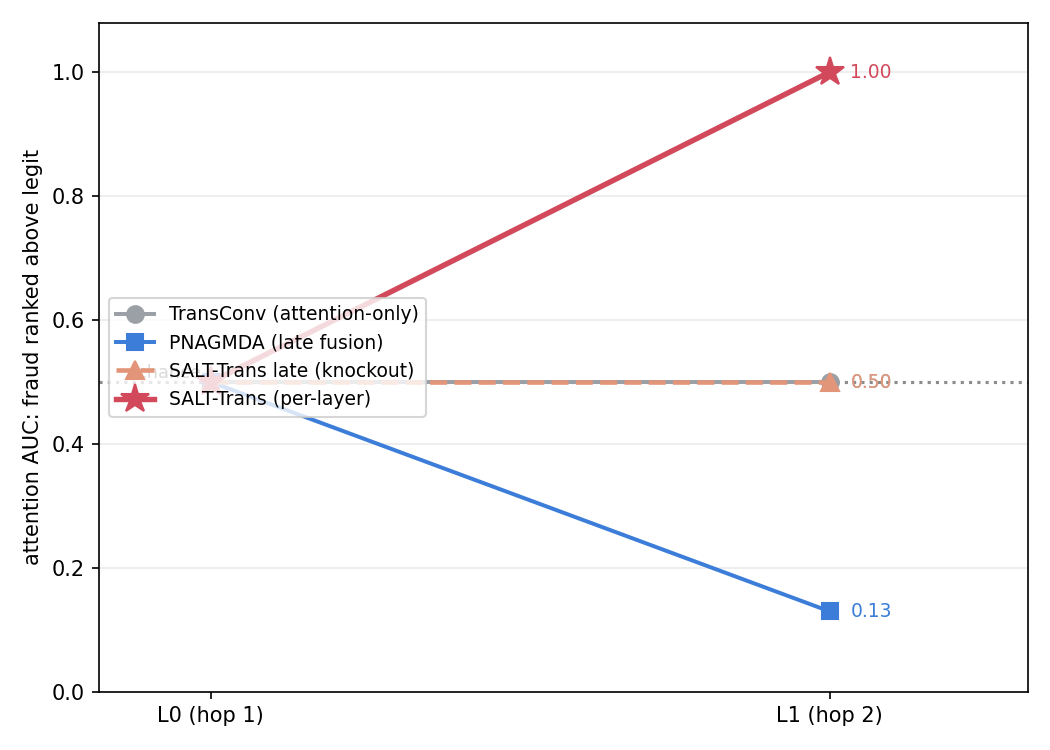}
\caption{Hop-resolved attention quality (AUC of fraud vs.\ legitimate mules ranked by
attention; $0.5$ = random). All models are at chance at the first hop (L0); only
per-layer fusion (SALT) lifts the second-hop (L1) attention onto the planted fraud.
Flipping SALT-Trans to prediction-time fusion (knockout) removes the lift.}
\label{fig:synth-attn}
\end{figure}

\end{document}